%% file: main.tex
\documentclass[10pt,twocolumn,letterpaper]{article}

\usepackage{wacv}
\usepackage{times}
\usepackage{epsfig}
\usepackage{graphicx}
\usepackage{amsmath}
\usepackage{amssymb}
\usepackage{booktabs}
\usepackage{microtype}
\usepackage{xcolor}
\usepackage{caption}
\usepackage[american]{babel}
\usepackage[pagebackref=true,breaklinks=true,letterpaper=true,colorlinks,bookmarks=false]{hyperref}

\wacvfinalcopy 

\def\wacvPaperID{220} 

\ifwacvfinal\fi
\begin{document}
\title{DAVID: Dual-Attentional Video Deblurring}

\author{Junru Wu$^{\dagger}$, Xiang Yu$^{\ddag}$, Ding Liu$^\ast$, Manmohan Chandraker$^{\S\ddag}$ and Zhangyang Wang$^\dagger$\\
$^{\dagger}$Texas A\&M University \\
$^{\ddag}$ NEC Laboratories America \\
$^{\S}$University of California, San Diego \\
$^{\ast}$ByteDance AI Lab\\
{\tt\small \{sandboxmaster,atlaswang\}@tamu.edu, \{xiangyu,manu\}@nec-labs.com, dingliu2@illinois.edu}\\}

\input{teaser}
\input{abstract}
\input{introduction2}

\input{related_work}
\input{method}

\input{experiment}

\input{conclusion}

{\small
\bibliographystyle{ieee}
\bibliography{egbib}
}

\end{document}


\title{DAVID: Dual-Attentional Video Deblurring\\ Supplementary Material}

\maketitle

\begin{table*}[htp!]
\centering
\begin{tabular}{ c & c & c & c & c & c}
    \toprule
    Layer & Kernel Size & Stride &Padding & Output Size & Skip Connection\\
    \midrule
    Input & 3x3 & 1x1 & 1x1 &  24 x H x W & Conv20\\
    \midrule
    Conv1 & 3x3 & 1x1 & 1x1 & 64 x H x W & \\
    Conv2 & 3x3 & 1x1 & 1x1 & 64 x H x W & Up4 \\
    Down1 & 3x3 & 1x1 & 1x1 & 64 x H/2 x W/2 & \\
    \hline
    Conv3 & 3x3 & 1x1 & 1x1 & 128 x H/2 x W/2 & \\
    Conv4 & 3x3 & 1x1 & 1x1 & 128 x H/2 x W/2 & Up3\\
    Down2 & 3x3 & 1x1 & 1x1 & 128 x H/4 x W/4 & \\
    \hline
    Conv5 & 3x3 & 1x1 & 1x1 & 256 x H/4 x W/4 & \\
    Conv6 & 3x3 & 1x1 & 1x1 & 256 x H/4 x W/4 & Up2\\
    Down3 & 3x3 & 1x1 & 1x1 & 256 x H/8 x W/8 & \\
    \hline
    Conv7 & 3x3 & 1x1 & 1x1 & 512 x H/8 x W/8 & \\
    Conv8 & 3x3 & 1x1 & 1x1 & 512 x H/8 x W/8 & Up1\\
    Down4 & 3x3 & 1x1 & 1x1 & 512 x H/16 x W/16 & \\
    \hline
    Conv9 & 3x3 & 1x1 & 1x1 & 1024 x H/16 x W/16 & \\
    Conv10 & 3x3 & 1x1 & 1x1 & 1024 x H/16 x W/16 & \\
    \hline
    Up1 & 3x3 & 1x1 & 1x1 & 1024 x H/8 x W/8 & Conv8\\
    Conv11 & 3x3 & 1x1 & 1x1 & 512 x H/8 x W/8 & \\
    Conv12 & 3x3 & 1x1 & 1x1 & 512 x H/8 x W/8 & \\
    \hline
    Up2 & 3x3 & 1x1 & 1x1 & 512 x H/4 x W/4 & Conv6\\
    Conv13 & 3x3 & 1x1 & 1x1 & 256 x H/4 x W/4 & \\
    Conv14 & 3x3 & 1x1 & 1x1 & 256 x H/4 x W/4 & \\
    \hline
    Up3 & 3x3 & 1x1 & 1x1 & 256 x H/2 x W/2 & Conv4\\
    Conv16 & 3x3 & 1x1 & 1x1 & 128 x H/2 x W/2 & \\
    Conv17 & 3x3 & 1x1 & 1x1 & 128 x H/2 x W/2  & \\
    \hline
    Up4 & 3x3 & 1x1 & 1x1 & 128 x H x W & Conv2\\
    Conv18 & 3x3 & 1x1 & 1x1 & 64 x H x W & \\
    Conv19 & 3x3 & 1x1 & 1x1 & 64 x H x W & \\
    \hline
    Conv20 & 3x3 & 1x1 & 1x1 & 3 x H x W & Input\\
    \bottomrule
  \end{tabular}
\caption{Detail Configuration of our Backbone Network}
\label{backbone_supp}
\end{table*}

\section{Video Deblurring Result}
We include more video deblurring results in the supplementary material zip file as ``$DAVID\_video\_deblur\_result.mp4$''
and hope that it can best illustrate the visual effects. We compare our method with DVD and DeblurGAN on the new Challenging DVD set. Highlighted from the video results, our DAVID model clearly generalizes better to handle more complicated blur. 

\section{Network Architecture}
As a reference for the detail network structure introduced in our main submission, we hereby introduce the whole structure in a bottom-up way. For our backbone network, we use a variant of U-Net where we add one more skip connection from input to the last convolution layer (Conv20) to further accelerate the convergence, as shown in Table \ref{backbone_supp}. Besides, every Up layer has a skip link connecting to its mirrored convolutional layer, \ie, Up1 is shown as input to Conv8, Up2 is serving as input to Conv6, etc.

We apply a shallower network structure for the proposed internal and external attention modules, which is a 12-layer U-Net structure as shown in Table~\ref{attention_backbone}. We use a shallower network here is because we believe feature map aggregation is a relatively easier task than the deblur task. 

Our DAVID overall framework consists of an external attention module  $\mathcal{A}_{e}$ build on top of 3 internal attention modules $\left\{\mathcal{A}_{i}\right\}_{i=1}^{3}$, where each internal attention module $\mathcal{A}_{i}$ has 4 backbone branches $\left\{D_{i}\right\}_{i=1}^{4}$. The internal attention modules $\mathcal{A}_{1}$, $\mathcal{A}_{2}$, $\mathcal{A}_{3}$ are trained  with samples from C-DVD-3, C-DVD-7, and C-DVD-11, respectively. Detailed overview of the proposed DAVID framework is shown in Figure \ref{overview}.

\begin{table*}[htp!]
\centering
\begin{tabular}{ c & c & c & c & c & c}
    \toprule
    Layer & Kernel Size & Stride &Padding & Output Size & Skip Connection\\
    \midrule
    Input & 3x3 & 1x1 & 1x1 &  24 x H x W &\\
    \midrule
    Conv1 & 3x3 & 1x1 & 1x1 & 64 x H x W & \\
    Conv2 & 3x3 & 1x1 & 1x1 & 64 x H x W & Up1 \\
    Down1 & 3x3 & 1x1 & 1x1 & 64 x H/2 x W/2 & \\
    \hline
    Conv3 & 3x3 & 1x1 & 1x1 & 128 x H/2 x W/2 & \\
    Conv4 & 3x3 & 1x1 & 1x1 & 128 x H/2 x W/2 & Up2\\
    Down2 & 3x3 & 1x1 & 1x1 & 128 x H/4 x W/4 & \\
    \hline
    Conv5 & 3x3 & 1x1 & 1x1 & 256 x H/4 x W/4 & \\
    Conv6 & 3x3 & 1x1 & 1x1 & 256 x H/4 x W/4 & \\
    \hline
    Up1 & 3x3 & 1x1 & 1x1 & 256 x H/2 x W/2 & Conv4\\
    Conv7 & 3x3 & 1x1 & 1x1 & 128 x H/2 x W/2 & \\
    Conv8 & 3x3 & 1x1 & 1x1 & 128 x H/2 x W/2  & \\
    \hline
    Up2 & 3x3 & 1x1 & 1x1 & 128 x H x W & Conv2\\
    Conv9 & 3x3 & 1x1 & 1x1 & 64 x H x W & \\
    Conv10 & 3x3 & 1x1 & 1x1 & (3 x N) x H x W\footnotemark & \\
    \bottomrule
  \end{tabular}
\caption{Detail Configuration of our Attention Branch}
\label{attention_backbone}
\end{table*}

\section{Multi-Phase Implementation Details}

\noindent\textbf{Phase 1: backbone branch pre-train.} The backbone branches $D_{1}, D_{2}, D_{3}, D_{4}$ are independently trained with taking in $1, 3, 5, 7$ frames, respectively. 
The convolutional layers are initialized with xavier initialization with standard deviation
\begin{align}
Var=\frac{2}{N_{in}+N_{out}}
\label{std}
\end{align}
where $N_{in}$ and $N_{out}$ is the number of input and output neurons, respectively. We apply the initial learning rate $1e^{-5}$ and multiply it by 0.96 every 100 epochs. We halt the training by observing on the validation set that performance is not increasing for consecutively 50 epochs.

\noindent\textbf{Phase 2: Internal attention module training.}

After Phase 1, we assemble each pre-trained backbone branch into the internal attention module.
To train each $\mathcal{A}_{i}$, we train the internal attention branch with a learning rate $1e^{-5}$ while fixing the weight of backbone branches. After 100 epochs, we unfreeze the backbone branches and fine-tune entire $A_{n,i}$ and $D_{i}$ with a smaller learning rate $2e^{-6}$ with another 200 epochs. We use xavier initialization with standard deviation as defined in Equation \ref{std} for all layers in the internal attention branch.

\noindent\textbf{Phase 3: Joint training with the external attention module:}

We load all the pre-trained internal attention modules from Phase 2 and fix the weights for each of them. We fine-tune the external attention branch with initial learning rate $1e^{-5}$. After 200 epochs, we unfreeze all internal attention modules and jointly fine-tune the entire DAVID model with a learning rate $2e^{-6}$ for another 200 epochs. We use xavier initialization with standard deviation defined in Equation \ref{std} for all layers in the external attention branch.

\section{Attention Map Visualization}

\begin{figure*}[htp!]
\centering
\begin{tabular}{cccc}
\includegraphics[width=0.34\textwidth]{./images/attention_map/alley/alley_00013_blur.png} &
\includegraphics[width=0.34\textwidth]{./images/attention_map/alley/alley_00013_outputs.png} \\ [-3pt]
 (1)  & (2)  \\
\end{tabular}
\begin{tabular}{cccc}
\includegraphics[width=0.022\textwidth]{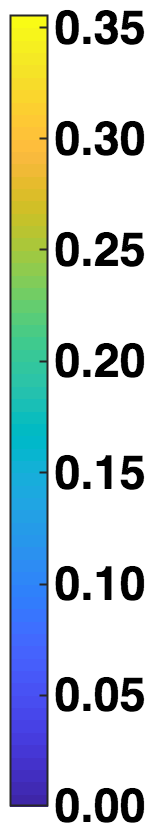} &
\includegraphics[width=0.22\textwidth]{./images/attention_map/alley/alley_00013_ext_1_att_heat.png} &
\includegraphics[width=0.22\textwidth]{./images/attention_map/alley/alley_00013_ext_2_att_heat.png} &
\includegraphics[width=0.22\textwidth]{./images/attention_map/alley/alley_00013_ext_3_att_heat.png} \\ [-3pt]
& (3)  & (4)  & (5) \\
\end{tabular}
\begin{tabular}{cccc}
\includegraphics[width=0.22\textwidth]{./images/attention_map/alley/alley_00013_ext_1_int_1_att_heat.png} &
\includegraphics[width=0.22\textwidth]{./images/attention_map/alley/alley_00013_ext_1_int_2_att_heat.png} &
\includegraphics[width=0.22\textwidth]{./images/attention_map/alley/alley_00013_ext_1_int_3_att_heat.png} &
\includegraphics[width=0.22\textwidth]{./images/attention_map/alley/alley_00013_ext_1_int_4_att_heat.png} \\ [-3pt]
(6)  & (7)  & (8) & (9) \\
\end{tabular}
\begin{tabular}{cccc}
\includegraphics[width=0.22\textwidth]{./images/attention_map/alley/alley_00013_ext_2_int_1_att_heat.png} &
\includegraphics[width=0.22\textwidth]{./images/attention_map/alley/alley_00013_ext_2_int_2_att_heat.png} &
\includegraphics[width=0.22\textwidth]{./images/attention_map/alley/alley_00013_ext_2_int_3_att_heat.png} &
\includegraphics[width=0.22\textwidth]{./images/attention_map/alley/alley_00013_ext_2_int_4_att_heat.png} \\ [-3pt]
(10)  & (11)  & (12) & (13) \\
\end{tabular}
\begin{tabular}{cccc}
\includegraphics[width=0.22\textwidth]{./images/attention_map/alley/alley_00013_ext_3_int_1_att_heat.png} &
\includegraphics[width=0.22\textwidth]{./images/attention_map/alley/alley_00013_ext_3_int_2_att_heat.png} &
\includegraphics[width=0.22\textwidth]{./images/attention_map/alley/alley_00013_ext_3_int_3_att_heat.png} &
\includegraphics[width=0.22\textwidth]{./images/attention_map/alley/alley_00013_ext_3_int_4_att_heat.png} \\ [-3pt]
(14)  & (15)  & (16) & (17) \\
\end{tabular}
\caption{ Visualization of external and internal attention maps on real blurry video ``alley'' in the qualitative testing set provided by \cite{su2017deep}}
\label{fig:attention_map_1}
\end{figure*}

\begin{figure*}[htp!]
\centering
\begin{tabular}{cccc}
\includegraphics[width=0.34\textwidth]{./images/attention_map/office/office_00040_blur.png} &
\includegraphics[width=0.34\textwidth]{./images/attention_map/office/office_00040_outputs.png} \\ [-3pt]
 (1)  & (2)  \\
\end{tabular}
\begin{tabular}{cccc}
\includegraphics[width=0.022\textwidth]{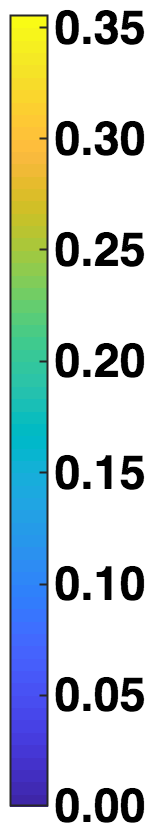} &
\includegraphics[width=0.22\textwidth]{./images/attention_map/office/office_00040_ext_1_att_heat.png} &
\includegraphics[width=0.22\textwidth]{./images/attention_map/office/office_00040_ext_2_att_heat.png} &
\includegraphics[width=0.22\textwidth]{./images/attention_map/office/office_00040_ext_3_att_heat.png} \\ [-3pt]
& (3)  & (4)  & (5) \\
\end{tabular}
\begin{tabular}{cccc}
\includegraphics[width=0.22\textwidth]{./images/attention_map/office/office_00040_ext_1_int_1_att_heat.png} &
\includegraphics[width=0.22\textwidth]{./images/attention_map/office/office_00040_ext_1_int_2_att_heat.png} &
\includegraphics[width=0.22\textwidth]{./images/attention_map/office/office_00040_ext_1_int_3_att_heat.png} &
\includegraphics[width=0.22\textwidth]{./images/attention_map/office/office_00040_ext_1_int_4_att_heat.png} \\ [-3pt]
(6)  & (7)  & (8) & (9) \\
\end{tabular}
\begin{tabular}{cccc}
\includegraphics[width=0.22\textwidth]{./images/attention_map/office/office_00040_ext_2_int_1_att_heat.png} &
\includegraphics[width=0.22\textwidth]{./images/attention_map/office/office_00040_ext_2_int_2_att_heat.png} &
\includegraphics[width=0.22\textwidth]{./images/attention_map/office/office_00040_ext_2_int_3_att_heat.png} &
\includegraphics[width=0.22\textwidth]{./images/attention_map/office/office_00040_ext_2_int_4_att_heat.png} \\ [-3pt]
(10)  & (11)  & (12) & (13) \\
\end{tabular}
\begin{tabular}{cccc}
\includegraphics[width=0.22\textwidth]{./images/attention_map/office/office_00040_ext_3_int_1_att_heat.png} &
\includegraphics[width=0.22\textwidth]{./images/attention_map/office/office_00040_ext_3_int_2_att_heat.png} &
\includegraphics[width=0.22\textwidth]{./images/attention_map/office/office_00040_ext_3_int_3_att_heat.png} &
\includegraphics[width=0.22\textwidth]{./images/attention_map/office/office_00040_ext_3_int_4_att_heat.png} \\ [-3pt]
(14)  & (15)  & (16) & (17) \\
\end{tabular}
\caption{ Visualization of external and internal attention maps on real blurry video ``office'' in the qualitative testing set provided by \cite{su2017deep}}
\label{fig:attention_map_2}
\end{figure*}

\begin{figure*}[htp!]
\centering
\begin{tabular}{cccc}
\includegraphics[width=0.34\textwidth]{./images/attention_map/boat/boat_00140_blur.png} &
\includegraphics[width=0.34\textwidth]{./images/attention_map/boat/boat_00140_outputs.png} \\ [-3pt]
 (1)  & (2)  \\
\end{tabular}
\begin{tabular}{cccc}
\includegraphics[width=0.022\textwidth]{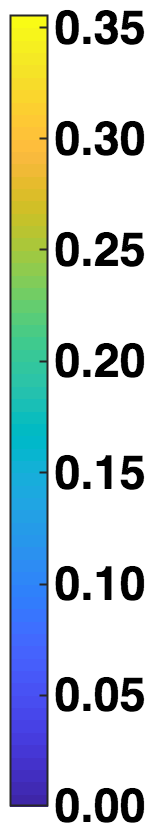} &
\includegraphics[width=0.22\textwidth]{./images/attention_map/boat/boat_00140_ext_1_att_heat.png} &
\includegraphics[width=0.22\textwidth]{./images/attention_map/boat/boat_00140_ext_2_att_heat.png} &
\includegraphics[width=0.22\textwidth]{./images/attention_map/boat/boat_00140_ext_3_att_heat.png} \\ [-3pt]
& (3)  & (4)  & (5) \\
\end{tabular}
\begin{tabular}{cccc}
\includegraphics[width=0.22\textwidth]{./images/attention_map/boat/boat_00140_ext_1_int_1_att_heat.png} &
\includegraphics[width=0.22\textwidth]{./images/attention_map/boat/boat_00140_ext_1_int_2_att_heat.png} &
\includegraphics[width=0.22\textwidth]{./images/attention_map/boat/boat_00140_ext_1_int_3_att_heat.png} &
\includegraphics[width=0.22\textwidth]{./images/attention_map/boat/boat_00140_ext_1_int_4_att_heat.png} \\ [-3pt]
(6)  & (7)  & (8) & (9) \\
\end{tabular}
\begin{tabular}{cccc}
\includegraphics[width=0.22\textwidth]{./images/attention_map/boat/boat_00140_ext_2_int_1_att_heat.png} &
\includegraphics[width=0.22\textwidth]{./images/attention_map/boat/boat_00140_ext_2_int_2_att_heat.png} &
\includegraphics[width=0.22\textwidth]{./images/attention_map/boat/boat_00140_ext_2_int_3_att_heat.png} &
\includegraphics[width=0.22\textwidth]{./images/attention_map/boat/boat_00140_ext_2_int_4_att_heat.png} \\ [-3pt]
(10)  & (11)  & (12) & (13) \\
\end{tabular}
\begin{tabular}{cccc}
\includegraphics[width=0.22\textwidth]{./images/attention_map/boat/boat_00140_ext_3_int_1_att_heat.png} &
\includegraphics[width=0.22\textwidth]{./images/attention_map/boat/boat_00140_ext_3_int_2_att_heat.png} &
\includegraphics[width=0.22\textwidth]{./images/attention_map/boat/boat_00140_ext_3_int_3_att_heat.png} &
\includegraphics[width=0.22\textwidth]{./images/attention_map/boat/boat_00140_ext_3_int_4_att_heat.png} \\ [-3pt]
(14)  & (15)  & (16) & (17) \\
\end{tabular}
\caption{ Visualization of external and internal attention maps on real blurry video ``boat'' in the qualitative testing set provided by \cite{su2017deep}}
\label{fig:attention_map_4}
\end{figure*}

We visualize more attention maps given an input frame in Figure~\ref{fig:attention_map_1}, Figure~\ref{fig:attention_map_2} and Figure~\ref{fig:attention_map_4}. Consistently, we observe the same trend as discussed in the paper experiment attention map visualization section. 

To reiterate, we see that the external attention maps (3) - (5) avoids constant degeneration and successfully highlights regional and structural information. When further looking across the three samples, we see that the magnitude for each attention map (from (3) to (5)) is different. For example, in Figure~\ref{fig:attention_map_1}, plot (4) overall has the most significant magnitude, plot (3) has the second most and plot (5) has the lowest. In Figure~\ref{fig:attention_map_2}, plot (5) shows the largest magnitude, plot (4) shows the second largest and plot(3) shows the least. This magnitude difference validates that our external attention module indeed adaptively selects the most informative channel according to the different blur level.

Further looking into plot (6) - (17), we observe the same trend that (6) - (9) show least magnitude which indicates that the DVD sample blur is not similar to the averaging 3 frame blur. Plot (10) - (13) show the attention maps with most structural information and plot (14) - (17) also shows moderate structural information. This suggests that DVD dataset samples are mostly with blur similar to averaging 7 and 11 frame blur. Moreover, it shows that the magnitude of (10) - (13) and (14) - (17) varies across different cases, which indicates that the internal attention modules adaptively select the most informative backbone branches with better deblur effects.





\section{Qualitative Results on Challenging DVD Dataset}

We show more visualization comparison of our David framework against \cite{su2017deep} in Figure~\ref{fig:visual_1} and Figure~\ref{fig:visual_2}. Inside each figure, we show two cases denoted as sub-figure (a) and sub-figure (b). The upper row shows the frame-level holistic deblur effect while the lower row shows the zoomed-in part from the upper row. Consistently, we observe that method of DVD (no-align) still remains large portion of blurry effects, \ie, Figure~\ref{fig:visual_1} sub-figure (b), the last row and second column, where the sunglasses are still blurry. Further comparing our DAVID trained on DVD original dataset and our proposed Challenging DVD (C-DVD) dataset, we see that DAVID (C-DVD) improves on top of DAVID (DVD) as less blurry effect is achieved. This suggests that our proposed C-DVD dataset contains more variant blur level samples which better reflects the real blur. We concludes that training on our proposed C-DVD dataset achieves a model with better generalization ability and better deblur performance.





\begin{figure*}[tbp]
\centering
\includegraphics[width=\textwidth]{./images/pipline_supp.png}
\caption{The detailed overview of the proposed DAVID framework. Each internal attention module is designed for a specific blur level. Each backbone branch in the same internal attention module works on a specific temporal scale.}
\label{overview}
\end{figure*}


\begin{figure*}[htp!]
\centering
(a)\\
\begin{tabular}{cccccc}
\includegraphics[width=0.23\textwidth]{./images/visual_compare/cdvd/1/1_00044_input_box.png} &
\includegraphics[width=0.23\textwidth]{./images/visual_compare/cdvd/1/1_00044_dvd_box.png} &
\includegraphics[width=0.23\textwidth]{./images/visual_compare/cdvd/1/1_00044_odvd_box.png} &
\includegraphics[width=0.23\textwidth]{./images/visual_compare/cdvd/1/1_00044_cdvd_box.png} \\ [-3pt]
\end{tabular}
\begin{tabular}{cccccc}
\includegraphics[width=0.23\textwidth]{./images/visual_compare/cdvd/1/1_00044_input_plot.png} &
\includegraphics[width=0.23\textwidth]{./images/visual_compare/cdvd/1/1_00044_dvd_plot.png} &
\includegraphics[width=0.23\textwidth]{./images/visual_compare/cdvd/1/1_00044_odvd_plot.png} &
\includegraphics[width=0.23\textwidth]{./images/visual_compare/cdvd/1/1_00044_cdvd_plot.png} \\ [-3pt]
\end{tabular}
(b)\\
\begin{tabular}{cccccc}
\includegraphics[width=0.23\textwidth]{./images/visual_compare/cdvd/4/4_00009_input_box.png} &
\includegraphics[width=0.23\textwidth]{./images/visual_compare/cdvd/4/4_00009_dvd_box.png} &
\includegraphics[width=0.23\textwidth]{./images/visual_compare/cdvd/4/4_00009_odvd_box.png} &
\includegraphics[width=0.23\textwidth]{./images/visual_compare/cdvd/4/4_00009_cdvd_box.png} \\ [-3pt]
\end{tabular}
\begin{tabular}{cccccc}
\includegraphics[width=0.23\textwidth]{./images/visual_compare/cdvd/4/4_00009_input_plot.png} &
\includegraphics[width=0.23\textwidth]{./images/visual_compare/cdvd/4/4_00009_dvd_plot.png} &
\includegraphics[width=0.23\textwidth]{./images/visual_compare/cdvd/4/4_00009_odvd_plot.png} &
\includegraphics[width=0.23\textwidth]{./images/visual_compare/cdvd/4/4_00009_cdvd_plot.png} \\[-3pt]
Blurry image & DVD (noalign) & DAVID (DVD) & DAVID (C-DVD) \\
\end{tabular}
\caption{Visual result comparison on our proposed Challenging DVD Datasets.}
\label{fig:visual_1}
\vspace{-3mm}
\end{figure*}

\begin{figure*}[htp!]
\centering
(a)\\
\begin{tabular}{cccccc}
\includegraphics[width=0.23\textwidth]{./images/visual_compare/cdvd/6/6_00028_input_box.png} &
\includegraphics[width=0.23\textwidth]{./images/visual_compare/cdvd/6/6_00028_dvd_box.png} &
\includegraphics[width=0.23\textwidth]{./images/visual_compare/cdvd/6/6_00028_odvd_box.png} &
\includegraphics[width=0.23\textwidth]{./images/visual_compare/cdvd/6/6_00028_cdvd_box.png} \\ [-3pt]
\end{tabular}
\begin{tabular}{cccccc}
\includegraphics[width=0.23\textwidth]{./images/visual_compare/cdvd/6/6_00028_input_plot.png} &
\includegraphics[width=0.23\textwidth]{./images/visual_compare/cdvd/6/6_00028_dvd_plot.png} &
\includegraphics[width=0.23\textwidth]{./images/visual_compare/cdvd/6/6_00028_odvd_plot.png} &
\includegraphics[width=0.23\textwidth]{./images/visual_compare/cdvd/6/6_00028_cdvd_plot.png} \\[-3pt]
\end{tabular}
(b) \\
\begin{tabular}{cccccc}
\includegraphics[width=0.23\textwidth]{./images/visual_compare/cdvd/9/9_00014_input_box.png} &
\includegraphics[width=0.23\textwidth]{./images/visual_compare/cdvd/9/9_00014_dvd_box.png} &
\includegraphics[width=0.23\textwidth]{./images/visual_compare/cdvd/9/9_00014_odvd_box.png} &
\includegraphics[width=0.23\textwidth]{./images/visual_compare/cdvd/9/9_00014_cdvd_box.png} \\ [-3pt]
\end{tabular}
\begin{tabular}{cccccc}
\includegraphics[width=0.23\textwidth]{./images/visual_compare/cdvd/9/9_00014_input_plot.png} &
\includegraphics[width=0.23\textwidth]{./images/visual_compare/cdvd/9/9_00014_dvd_plot.png} &
\includegraphics[width=0.23\textwidth]{./images/visual_compare/cdvd/9/9_00014_odvd_plot.png} &
\includegraphics[width=0.23\textwidth]{./images/visual_compare/cdvd/9/9_00014_cdvd_plot.png} \\[-3pt]
Blurry image & DVD (noalign) & DAVID (DVD) & DAVID (C-DVD) \\
\end{tabular}
\caption{Visual result comparison on our proposed Challenging DVD Datasets.}
\label{fig:visual_2}
\vspace{-3mm}
\end{figure*}

{\small
\bibliographystyle{ieee}
\bibliography{egbib}
}

%% file: teaser.tex
\twocolumn[{
\renewcommand\twocolumn[1][]{#1}
\maketitle
\begin{center}
\centering
\setlength{\tabcolsep}{0.3mm}
\renewcommand{\arraystretch}{0.25}
\vspace{-8mm}
\begin{tabular}{cccc}
\includegraphics[width=4.3cm]{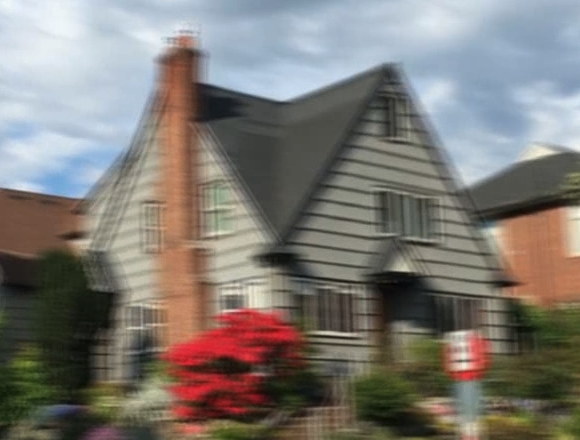} &
\includegraphics[width=4.3cm]{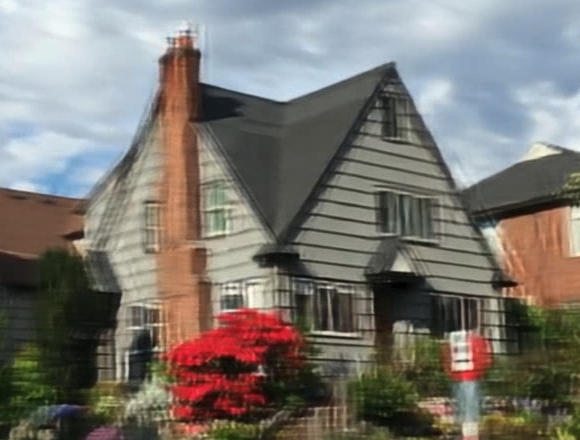} &
\includegraphics[width=4.3cm]{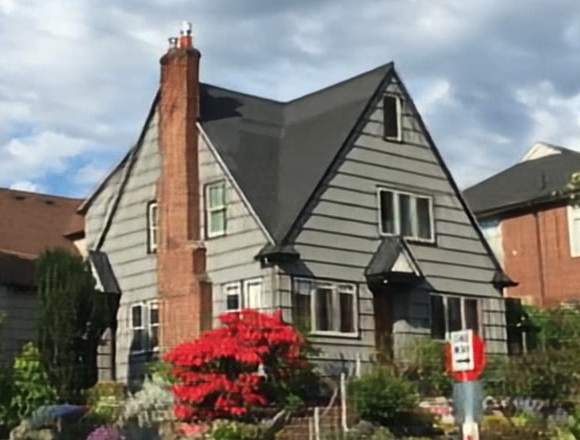} &
\includegraphics[width=4.3cm]{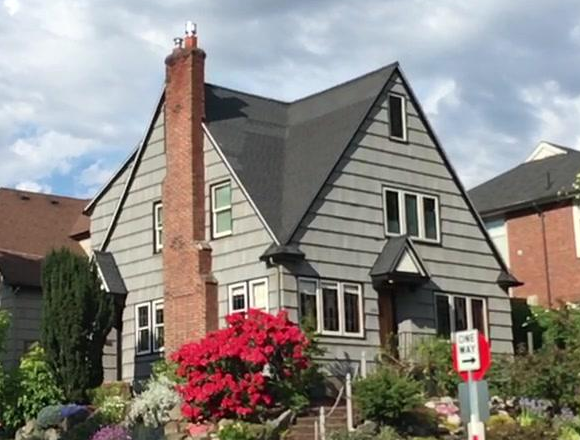}
\end{tabular}
\begin{tabular}{cccc}
\includegraphics[width=4.3cm]{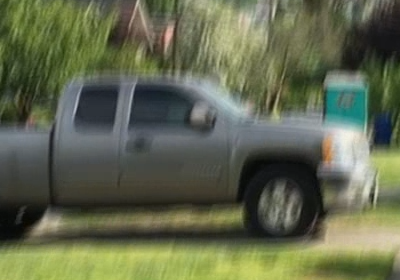} &
\includegraphics[width=4.3cm]{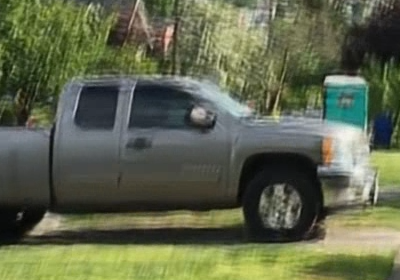} &
\includegraphics[width=4.3cm]{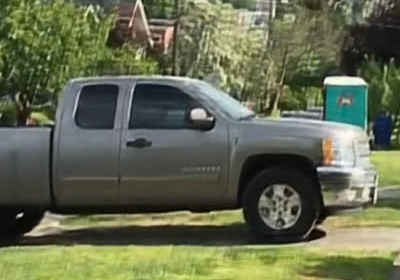} &
\includegraphics[width=4.3cm]{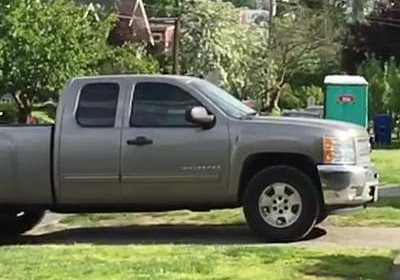} \\ [1pt]
Input & DVD (noalign) & DAVID (Ours) & Groud Truth \\
\end{tabular}
\vspace{-3mm}
\captionof{figure}{Visual comparison on Challenging DVD Dataset with DVD (noalign)~\cite{su2017deep}.}
\label{fig:teaser}
\vspace{-1.5mm}
\end{center}}]

%% file: abstract.tex
\begin{abstract}
\vspace{-4mm}

Blind video deblurring restores sharp frames from a blurry sequence without any prior. It is a challenging task because the blur due to camera shake, object movement and defocusing is heterogeneous in both temporal and spatial dimensions. Traditional methods train on datasets synthesized with a single level of blur, and thus do not generalize well across levels of blurriness. To address this challenge, we propose a dual attention mechanism to dynamically aggregate temporal cues for deblurring with an end-to-end trainable network structure. Specifically, an internal attention module adaptively selects the optimal temporal scales for restoring the sharp center frame. An external attention module adaptively aggregates and refines multiple sharp frame estimates, from several internal attention modules designed for different blur levels. To train and evaluate on more diverse blur severity levels, we propose a Challenging DVD dataset generated from the raw DVD video set by pooling frames with different temporal windows. Our framework achieves consistently better performance on this more challenging dataset while obtaining strongly competitive results on the original DVD benchmark. Extensive ablative studies and qualitative visualizations further demonstrate the advantage of our method in handling real video blur. 
\end{abstract}

%% file: introduction2.tex
\section{Introduction}
Mobile phones, high internet bandwidths and social media have led to a recent spurt in video acquisition and sharing. However, videos of dynamic scenes, or those captured through hand-held devices, often display spatially and temporally varying blur patterns. The source of blur can be camera shake, depth variation, object motions or a combination of them, which manifest in complex patterns such as jittering, jumping or ghosting artifacts. Restoration of spatial structure and image sharpness is an ill-posed problem, for which single-image deblurring methods have been proposed relying on statistical \cite{freeman2009understanding,hyun2014segmentation} or learned priors \cite{nah2017deep,schuler2016learning,tao2018scale,xu2017learning, kupyn2019deblurgan}. But video deblurring remains relatively less-studied, since effectively and efficiently modeling the inherent temporal dynamics among consecutive video frames is challenging.

Recent works on video deblurring usually work on a fixed temporal scale. For example, a fixed number of blurry frames are stacked as inputs to the network in \cite{su2017deep,kim2017online}. However, the spatio-temporal scale of blurs can vastly vary. Camera shake usually leads to short-term, spatially uniform and temporally uncorrelated blurs \cite{whyte2012non,su2017deep}, while object motion causes long-term, spatially localized and temporally smooth blurs \cite{pan2016soft}. Such heterogeneity in blur patterns is unlikely to be well represented by a single model in the above approaches, and calls for the selection or modulation of temporal scales.

This paper proposes a Dual-Attentional VIdeo Deblurring (\textbf{DAVID}) framework, aiming to simultaneously model temporal dynamics among consecutive frames and handle various levels of spatially heterogeneous blur in real videos. We start by constructing a compact backbone module based on a U-Net variant \cite{ronneberger2015u}, to infer the sharp center frame from a number of consecutive blurry frames. Using this network module as a basic building block, we propose two levels of attention modules to hierarchically reason about blur compositions and levels in both temporal and spatial dimensions. 

\begin{figure*}[tbp]
\vspace{-2em}
\centering
\includegraphics[width=\textwidth]{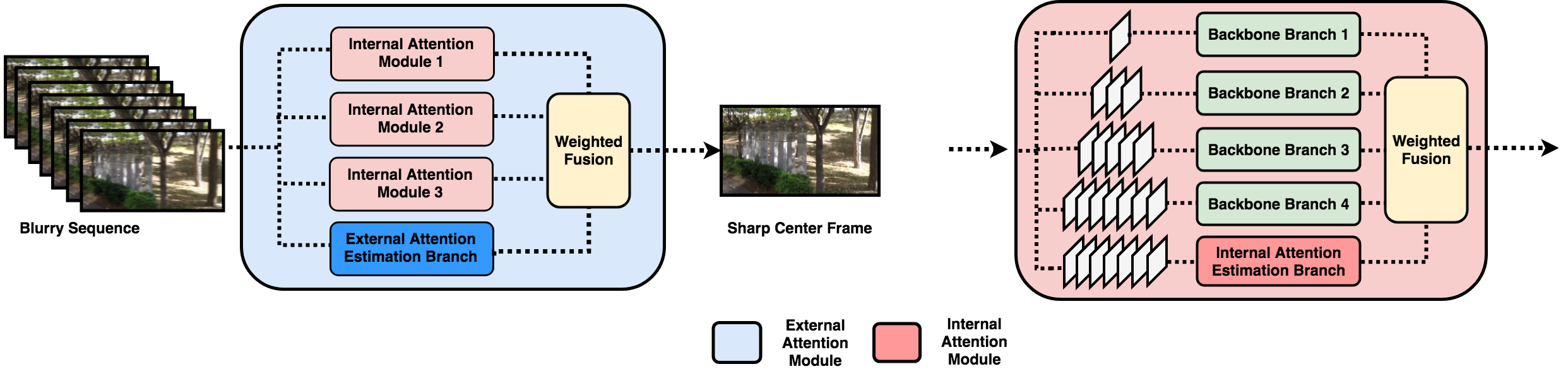}
\caption{The overview of the proposed DAVID framework. Left: the structure of the external attention module. Right: the structure of the internal attention module. Each internal attention module is designed for a specific blur level. Each backbone branch in the same internal attention module works on a specific temporal scale.}
\label{overview}
\end{figure*}

We determine the optimal scale of temporal dependency through an \textbf{internal attention} module that takes as input several consecutive blurry frames, while multiple backbone branches work on different temporal scales to generate different sharp frame estimates. The resultant sharp frame estimates are adaptively aggregated according to the attention map, which is inferred by another branch in this module. Further, an \textbf{external attention} module handles various levels of spatially heterogeneous blur by determining how to adaptively aggregate the outputs of multiple internal attention modules, each of which is designed for a specific blur level. The architecture is illustrated in Figure \ref{overview}. To the best of our knowledge, this is the first work in video deblurring that exploits hierarchical attention.

Our proposed method achieves consistently better PSNR compared to others on public datasets such as DVD. Further, we create a new video deblurring dataset, called \textit{Challenging Deep Video Deblurring} (Challenging DVD), to introduce stronger blur variations by synthesis. Compared to the original Deep Video Deblurring (DVD) dataset proposed in \cite{su2017deep}, we witness performance drops for state-of-the-art methods when applied to Challenging DVD. Extensive ablation studies and visual results on real blur videos further indicate that \textbf{DAVID} can adaptively deal with different levels of blurriness, leading to perceptible improvement in quality.

In summary, our main contributions are:
\begin{itemize}
\vspace{-0.5em}
\item A novel Dual-Attentional Video Deblurring (\textbf{DAVID}) framework, which jointly takes into account heterogeneous blur information in temporal and spatial dimensions for video deblurring, with an external attention and several internal attention modules. 
\vspace{-0.5em}
\item A new video deblurring dataset, Challenging Deep Video Deblurring (Challenging DVD), that introduces stronger blur composition and level variations.
\vspace{-0.5em}
\item Extensive experiments on both Challenging and original DVD datasets, to demonstrate that DAVID achieves consistently better or competitive results, respectively.
\end{itemize}

%% file: related_work.tex
\section{Related Work}

\noindent\textbf{Image Deblurring:} Image deblurring can be traced back to the traditional methods using different types of priors, such as total variation (TV) \cite{hyun2014segmentation}, sparsity \cite{freeman2009understanding}, heavy-tailed gradient prior \cite{shan2008high}, and $l_0$-norm gradient prior \cite{xu2013unnatural}. Many early approaches \cite{kohler2012recording,krishnan2011blind,harmeling2010space,hirsch2011fast} proposed to estimate the blur kernel and then apply deconvolution, in which the estimated kernel quality significantly impacted the result. \cite{pan2016blind} presented a simple and effective dark channel prior. \cite{ren2016image} showed that a simple low-rank model significantly reduces blur even without using any kernel information. Segmentation information \cite{hyun2013dynamic} was also investigated as an accompanying cue for motion.


In recently years, CNN-based image deblurring methods have achieved success \cite{xu2014deep}. \cite{sun2015learning} proposed to directly estimate motion blur fields by CNN. \cite{schuler2016learning} adopted a coarse-to-fine manner to stack multiple CNNs to analyze the blur formation. \cite{xu2018motion} proposed to estimated motion blur kernel using deep learning. 
\cite{nah2017deep} presented multi-scale
loss function that mimics conventional coarse-to-fine
approaches in training a multi-scale deblurring CNN. \cite{kupyn2017deblurgan} introduced generative adversarial networks (GANs) to obtain sharp and realistic-looking images.
\cite{tao2018scale} introduced a scale-recurrent network to exploit the multi-scale spatial information. Recent works also shows that restoration of blurry image can be used to facilitate high-level task such as object detection \cite{yuan2019ug}, image classification\cite{vidalmata2019bridging} and image segmentation \cite{wang2019segmentation}.



\noindent\textbf{Video Deblurring:} 
In the video case, temporal variations critically determine the blur effect, and provide additional clues for deblurring. Early video deblurring methods \cite{matsushita2006full,cho2012video} compensated sharp details of the current frame by nearby frames, via patch matching, motion flow and frame alignment. They however failed easily when dealing with large movements. 
\cite{delbracio2015hand} applied optical flow to warp nearby frames and fused them in the Fourier domain, but suffered from unreliable flow estimation when occlusions or outliers are present. \cite{pan2017simultaneous} proposed to simultaneously
deblur stereo videos and estimate the scene flow, where the motion cues from the scene flow and blur information can complement each other.

In deep learning, video deblurring has so far received relatively limited attentions.
\cite{su2017deep} proposed DeBlurNet (DBN) on accumulating information across frames, where neighboring frames were stacked as inputs to predict the clean central frame. \cite{kim2017online} introduced a spatio-temporal recurrent network that adaptively enforced temporal consistency between consecutive frames. 
Segmentation information has also been incorporated with video deblurring when available \cite{ren2017video}.



\noindent\textbf{Attention Model:} Attention in general serves as a learnable guidance, to re-allocate available processing resources towards the most informative input components. It has shown promise in language translation \cite{vaswani2017attention}, object recognition \cite{ba2014multiple}, image generation \cite{selfattention} and person re-id \cite{chen2019abd}, meta-learning \cite{cao2019learning}. For video-based applications, \cite{nan2017} proposed a temporal attention network to aggregate video frames for face recognition. \cite{liu2017robust} developed attention-based temporal modulation for video super-resolution, where the predictions made within different temporal windows were adaptively fused with a learned pixel-level attention map. We also note that a few others \cite{wang2016hierarchical,kosiorek2017hierarchical} discussed the concept of hierarchical attention, but with completely different contexts and motivations from ours. 

%% file: method.tex
\section{The Proposed Model: DAVID}

\subsection{Network Design}

The overall architecture of our DAVID model is illustrated in Figure \ref{overview}. Taking a set of $2N - 1$ consecutive blurry video frames, DAVID aims to output the  center sharp frame, \ie, the $N$-th frame. Each internal attention module determines the optimal temporal scale for removing blur, via adaptively aggregating multi-scale temporal information from several \textit{backbone branches}. The external attention then estimates the spatially global blur level, and make soft assignments for several \textit{internal attention modules}, each of which is dedicated for a specific blur level. Finally, their outputs are fused adaptively under the guidance of external attention, to generate the sharp frame.

\begin{figure*}[htbp]
\vspace{-4em}
\centering
\includegraphics[width=0.95\textwidth]{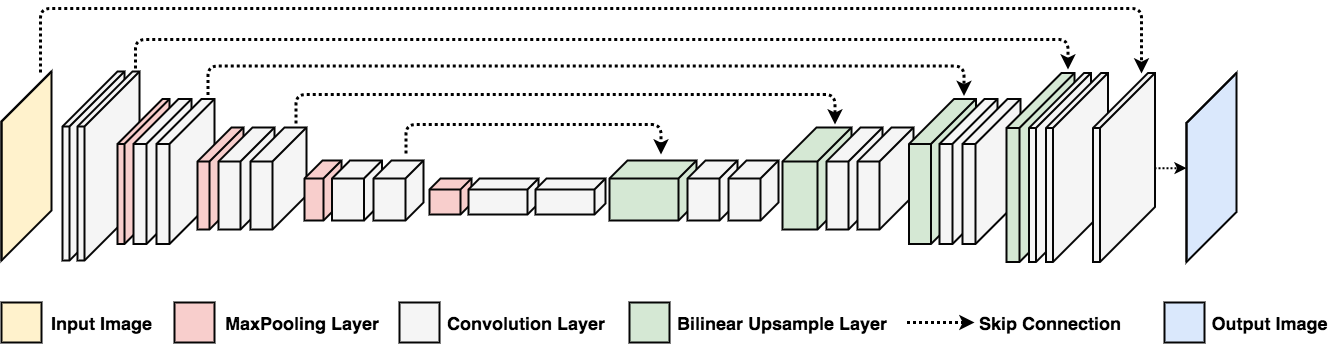}
\caption{A proposed backbone branch for DAVID.}
\label{pipeline}
\vspace{-0.5em}
\end{figure*}

\vspace{-1em}
\paragraph{Backbone Branch:} 
Our backbone branch adopts an encoder-decoder U-Net, which is very popular for many image restoration tasks \cite{liu2017image,mao2016image,tao2017detail,tao2018scale}, as detailed in Figure \ref{pipeline}. The encoder consists of five blocks. Each block contains two consecutive convolutional layers, followed by a max-pooling layer, which downsamples the feature maps by half. The decoder is formulated in a mirroring way, with five blocks. Each one contains a bilinear upsampling layer followed by two convolutional layers. The feature maps after each block is upsampled by a factor of two and thus the final output achieves the same size as the input. Skip connections are adopted to aggregate the feature maps from the encoder to the decoder on each spatial scale, to better utilizes features from multiple scales.

\begin{figure}[htbp]
\centering
\includegraphics[width=0.48\textwidth]{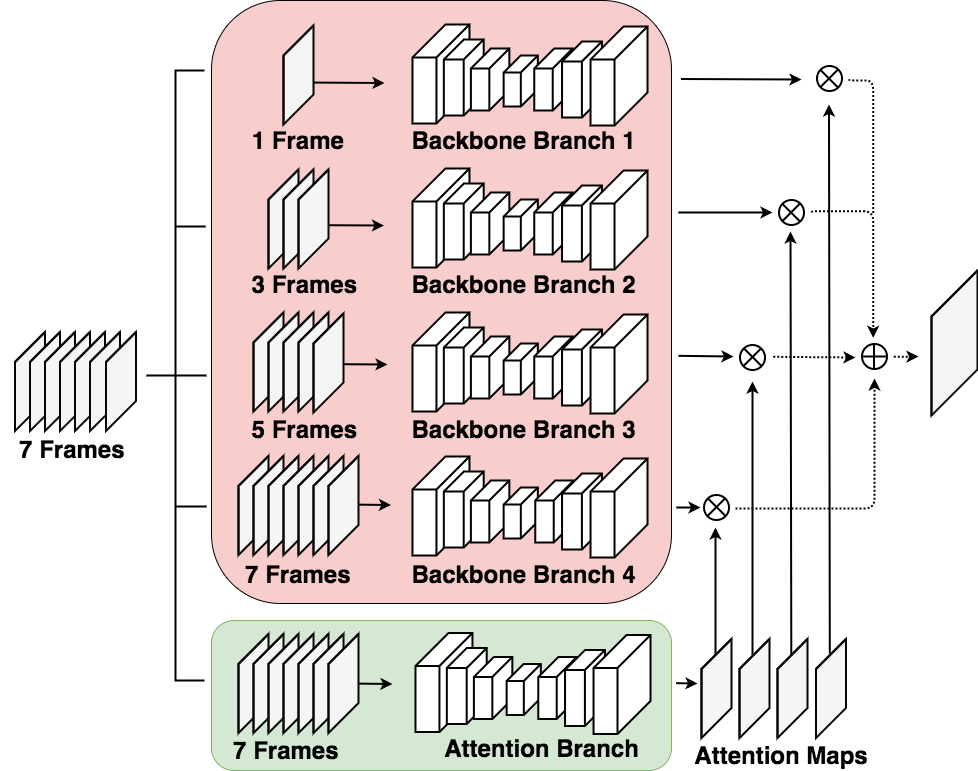}
\vspace{-0.5em}
\caption{An internal-attentional module for DAVID.}
\label{internal_attention}
\vspace{-0.5em}
\end{figure}

%
%


%
As the building block of DAVID, we carefully fine-tune the backbone to obtain remarkable effectiveness: as shown in experiments later, a single backbone branch is already able to outperform \cite{su2017deep} on the DVD dataset. Its performance will be further boosted as the temporal information is added.  
\vspace{-1em}
\paragraph{Internal Attentional Module:}
As shown in Figure~\ref{internal_attention}, an internal attention module $\mathcal{A}_{n}$ is designed to coordinate a number of backbone branches.
Each backbone branch handles a different temporal scale: The $D_{i}$ backbone branch takes $2i + 1$ consecutive blurry input frames (centered at the current frame) to predict the center sharp estimate. 
As a result, the convolutional filters in the first layer are customized to have $(2i + 1) \times r$ channels, where $r$ denotes the channel number of each input frame.

Besides the backbone branches, we design an \textit{internal attention estimation branch} to learn the selectivity on different temporal scales according to blur information, by predicting pixel-level aggregation weights for each branch's output.
In practice, the internal attention estimation branch takes all $2N-1$ frames as inputs, and outputs the pixel-level weight maps on all $N$ possible temporal scales. Considering the computation cost and efficiency, we adopt a shallower  architecture of two downsampling blocks for the  estimation branch. 
Eventually, each backbone branch's output $D_{i}$ is pixel-wisely multiplied with its corresponding weight map from the internal attention estimation branch, and these products are summed up to form the final output of the internal attention module. This operation is expressed in Equation~\ref{eq:internal_att}: 
\begin{align}
\vspace{-2mm}
\hat{I}_{n}=\sum_{i=1}^{N}\mathcal{A}_{n,i} \otimes D_{i}
\label{eq:internal_att}
\vspace{-2mm}
\end{align}
$\mathcal{A}_{n,i}$ denotes the internal attention map for backbone branch $i$. $D_{i}$ is the output from backbone branch $i$. $\otimes$ represents pixel-wise multiplication.
$N$ is the total number of backbone branches and
$\hat{I}_{n}$ is the recovered output from internal attention module.






\vspace{-1em}
\paragraph{External Attentional Module:}
In DAVID, each internal attention module focuses on one global background blur level. %
The assumption on the blur level further needs to be aggregated in a probabilistic way so that we can better model the spatially variant blur effect. Thus, an external attention mechanism $\mathcal{A}_{e}$ is proposed on top of the internal attention modules $\mathcal{A}_{n}$.
Similar to the internal attention module, we design an \textit{external attention branch} to predict a pixel-level weight map for each internal attention module, with all $2N-1$ frames as its input. Its implementation also refers to a backbone branch architecture with only two
downsampling blocks. The final deblurring result by DAVID is the pixel-wise weighted summation result of all internal attention module outputs, as in Equation~\ref{eq:external_att}.
\begin{align}
\vspace{-2mm}
\hat{I}_{e}=\sum_{j=1}^{M}\mathcal{A}_{e,j} \otimes \hat{I}_{n,j}
\label{eq:external_att}
\vspace{-2mm}
\end{align}
$\mathcal{A}_{e,j}$ is the external attention map for the $j$-th internal attention module output $\hat{I}_{n,j}$.
$M$ is the total number of internal modules and $\hat{I}_{e}$ the final output from the external module.

\subsection{Challenging DVD Dataset}

To train data-driven video deblurring models, a large number of pairs of blurry and sharp frames are needed. Although it is possible to use a beam-splitter and multiple cameras to build a special capturing system,  this setup would be challenging to construct robustly, and can face many calibration issues. \cite{su2017deep} proposed to collect real-world sharp videos at very high frame rate, and synthetically create blurred ones by averaging consecutive short-exposure images to approximate a longer exposure \cite{telleen2007synthetic}. The authors collected 71 videos at 240 frames per seconds (fps), with 3-5s average length.  In order to simulate realistic blurs at 30fps, they subsampled every eighth frame to create the 30fps ground truth sharp video, while averaging 7 neighboring frames (centered at the corresponding ground truth frame) to generate the blurred video at 30fps. The video deblurring model is then trained to recover the 30fps ground truths from the 30fps blurred versions. We refer to this resulting dataset as the Deep Video Deblurring (DVD) dataset, which has been the most popular benchmark for video deblurring algorithms.

To mimic more challenging real-world blurs, we create a new \textbf{Challenging DVD dataset}, aiming to cover a wider spectrum of blur variations. This is achieved by introducing more stochasticity during blurry video synthesis. Using the same high fps video of DVD, we randomly choose every 3, 7, 11 or 15 frames to average into the blurry frame,\footnote{We choose 3, 7, 11, 15 here, since we hope that they can represent several different blur levels from light to severe. Those choices are empirical, in no way unique, and may be further tuned for better practical performance.} instead of only averaging 7 frames as in the DVD dataset.\footnote{\cite{su2017deep} suggest to use optical flow to interpolate additional frames in order to smoothen the averaging. However, we observe that the classical optical flow often introduce visible ``ghost'' artifacts in the synthetic blurry frames. We thus use a neural network based frame interpolation model \cite{niklaus2017video} to replace that, observing improved blur quality.} We dubbed them C-DVD-3, C-DVD-7, C-DVD-11 and C-DVD-15, respectively. The modification leads to the new Challenging DVD dataset that contains dynamic, and often more severe (when 11 or 15 frames averaged) blurs. Experiments manifest that almost all video deblurring methods witness performance drops when applied to Challenging DVD, compared to performance on DVD. The value of Challenging DVD dataset is justified later by our model's generalization evaluation on real-world blurry videos when trained on it.



%% file: experiment.tex
\section{Experiments}
\vspace{-0.5em}
In this section, we firstly explain the detail implementation and the training protocols. Then, we compare DAVID to several video-based state-of-the-arts on the Challenging DVD dataset, to highlight our method's effectiveness on different levels of blur. The performance of DAVID on the original DVD dataset is also reported for reference comparison. We then investigate the ablative functions for each of our proposed modules, \ie, single back bones, internal attention modules.
Visualization of our attention modules is provided for further insightful analysis. Finally, we display qualitative results of DAVID on real blurry videos and compare to the state-of-the-arts.

\subsection{Multi-Phase Training for DAVID}
To train our Dual-Attention Video Deblurring (\textbf{DAVID}) framework, good initializations are needed for both backbone branches and internal/external attention modules. We propose a multi-phase strategy to smoothly conduct training.

\textbf{Phase 1: Pre-train each backbone branch:}
We first pre-train each backbone branch $D_{i}$. Specifically, the backbone branches $D_{1}$, $D_{2}$, $D_{3}$, $D_{4}$ are independently trained to take in $1$, $3$, $5$, $7$ frames, respectively. 

\textbf{Phase 2: Pre-train the internal attention module:}
After Phase 1, we assemble each pre-trained backbone branch into the internal attention module. The internal attention modules $\mathcal{A}_{n,1}$, $\mathcal{A}_{n,2}$, $\mathcal{A}_{n,3}$ are trained separately, with samples from C-DVD-3, C-DVD-7, and C-DVD-11, respectively, where each represents a different level of blur that is temporally pooled from 3, 7, 11 frames. 
To train each $\mathcal{A}_{i}$, we train the internal attention branch with a learning rate $1e^{-5}$ while fixing the weight of backbone branches. After 100 epochs, we unfreeze the backbone branches and fine-tune entire $A_{n,i}$ and $D_{i}$ with a smaller learning rate $2e^{-6}$ until convergence. In this way, we enforce each $A_{n,i}$ to focus on a specific level of blur. The learning rate and hyper-parameters are grid searched based on a validation set.

\textbf{Phase 3: Joint training with the external attention module:}
We load all the pre-trained internal attention modules from Phase 2 and fix the weights for each of them. We fine-tune the external attention branch with initial learning rate $1e^{-5}$. After 200 epochs, we unfreeze the attention branch in each of the internal attention module, jointly fine-tune the internal/external attention branches for 200 epoch with a learning rate $2e^{-6}$. Finally, we unfreeze all the backbone branches in each of the internal attention module and jointly fine-tune the entire DAVID model with a learning rate $2e^{-6}$ for another 200 epochs.

\noindent\textbf{Implementation Details}
We implement our backbone branch with PyTorch. By default, our model has 3 internal attention modules, each consists of 4 single backbone branches, that take 1, 3, 5, 7 consecutive input frames, respectively. Please refer to the detail network structure design in the supplementary material.
During training, we apply random cropping, random horizontal and vertical flipping as data augmentation. Batch size is set to $16$ across all the training phases. We adopt Adam~\cite{kingma2014adam} solver with $L_{2}$ loss. Initial learning rate is set $0.0001$. 
The momentum and weight decay are fixed as $0.9$ and $0$, respectively.

\begin{table*}[htp!]
\vspace{-2em}
\centering
\begin{tabular}{c | c  c  c  c  c  c  c  c  c  c | c }
    \toprule
    Methods & 1 & 2 & 3 & 4 & 5 & 6 & 7 & 8 & 9 & 10 & Average \\
    \midrule
    WFA\cite{delbracio2015burst} & $25.16$ & $29.42$ & $32.73$ & $23.88$ & $22.91$ & $29.77	$ & $26.67$ & $25.36$ & $31.10$ & $23.68$ & $26.34$ \\
    DVD (no align)\cite{su2017deep} & $27.37$ & $32.00$ & $34.93$ & $25.88$ & $25.02$ & $31.68$ & $28.87$ & $27.47$ & $33.23$ & $25.79$ & $28.52$ \\
    \hline
    Single (C-DVD-3) & $26.91$ & $32.77$ & $37.44$ & $24.85$ & $24.93$ & $31.43$ & $28.51$ & $26.76$ & $32.88$ & $24.61$ & $28.52$\\
    Single (C-DVD-7) & $29.69$ & $35.19$ & $39.11$ & $27.13$ & $26.68$ & $34.10	$ & $31.05$ & $29.06$ & $36.60$ & $27.65$ & $31.03$ \\
    Single (C-DVD-11) & $29.72$ & $35.63$ & $38.99$ & $28.15$ & $26.90$ & $34.28	$ & $31.19$ & $29.33$ & $36.55$ & $28.44$ & $31.31$ \\
    Internal Att (C-DVD-3) & $27.34$ & $33.11$ & $37.74$ & $25.11$ & $25.20$ & $31.84$ & $28.95$ & $27.05$ & $33.29$ & $24.85$ & $28.86$ \\
    Internal Att (C-DVD-7) & $30.05$ & $35.57$ & $39.25$ & $27.45$ & $26.95$ & $34.41$ & $31.37$ & $29.27$ & $36.85$ & $27.93$ &	$31.32$ \\
    Internal Att (C-DVD-11) & $30.17$ & $35.95$ & $39.03$ & $28.42$ & $27.20$ & $34.48$ & $31.49$ & $29.38$ & $36.73$ & $28.67$ & $31.56$ \\
    Dual Att (C-DVD-3-7-11) & $\textbf{30.67}$ & $\textbf{36.23}$ & $\textbf{39.26}$ & $\textbf{28.58}$ & $\textbf{27.41}$ & $\textbf{34.82}$ & $\textbf{31.73}$ & $\textbf{29.50}$ & $\textbf{37.35}$ & $\textbf{28.94}$ & $\textbf{31.84}$\\
    \bottomrule
  \end{tabular}
  \vspace{-2mm}
\caption{PSNR results on the full Challenging DVD testing set. Best results are shown in bold.}
\label{Challenging_DVD}
\end{table*}

\begin{table}[btp!]
\centering
\begin{tabular}{ @{\hskip 1mm}c| @{\hskip 1mm}c | @{\hskip 1mm}c |@{\hskip 1mm}  c |@{\hskip 1mm}  c |@{\hskip 1mm} c }
    \hline
     stacked frames & 1 & 3 & 5 & 7 & 9 \\
    \hline
    C-DVD-11 & $28.91$ & $29.80$ & $29.81$ & $\textbf{29.84}$ & $29.82$\\
    \hline
  \end{tabular}
\caption{Effect of multiple stacked frames as the input for the single backbone branch, trained on C-DVD-11 subset, tested on C-DVD-11 testing set.}
\vspace{-4mm}
\label{ablation_depth}
\end{table}

\begin{figure*}[htp!]
\centering
\begin{tabular}{@{}cc@{}}
\includegraphics[width=0.34\textwidth]{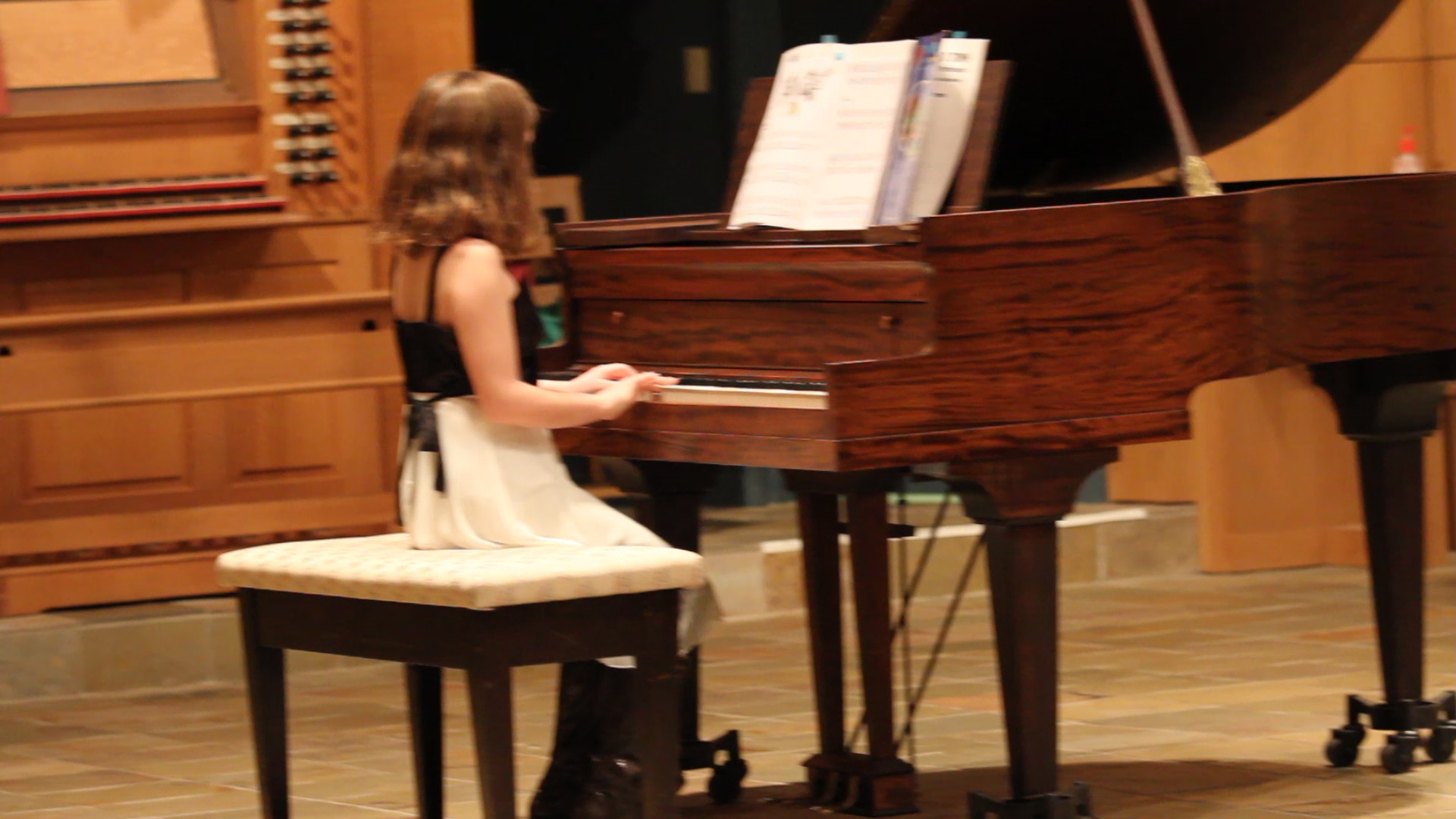} &
\includegraphics[width=0.34\textwidth]{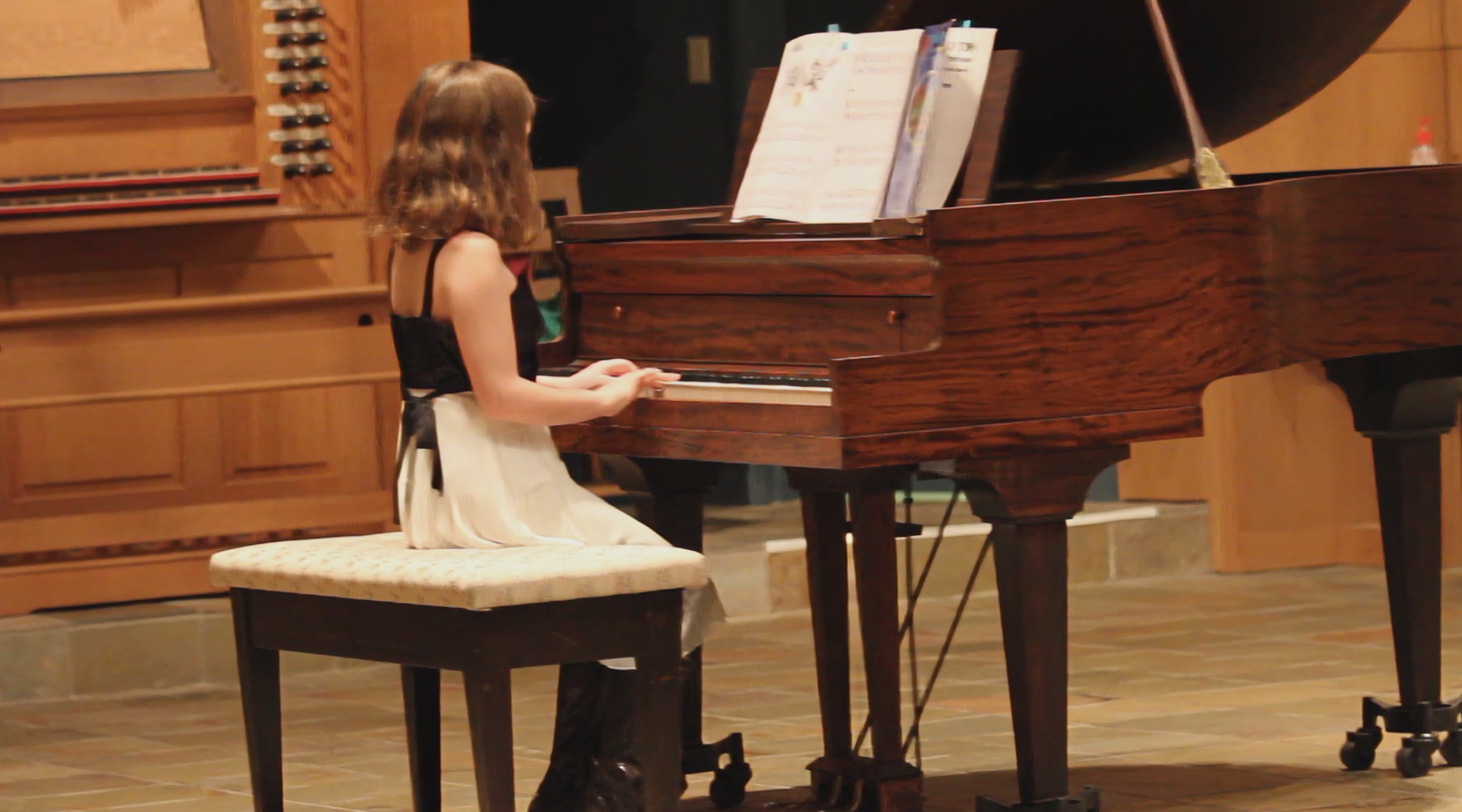} \\ [-3pt]
 (1)  & (2)  \\
\end{tabular}
\begin{tabular}{@{}ccccc@{}}
\includegraphics[width=0.025\textwidth]{./images/attention_map/piano/bar_new.png} &
\includegraphics[width=0.23\textwidth]{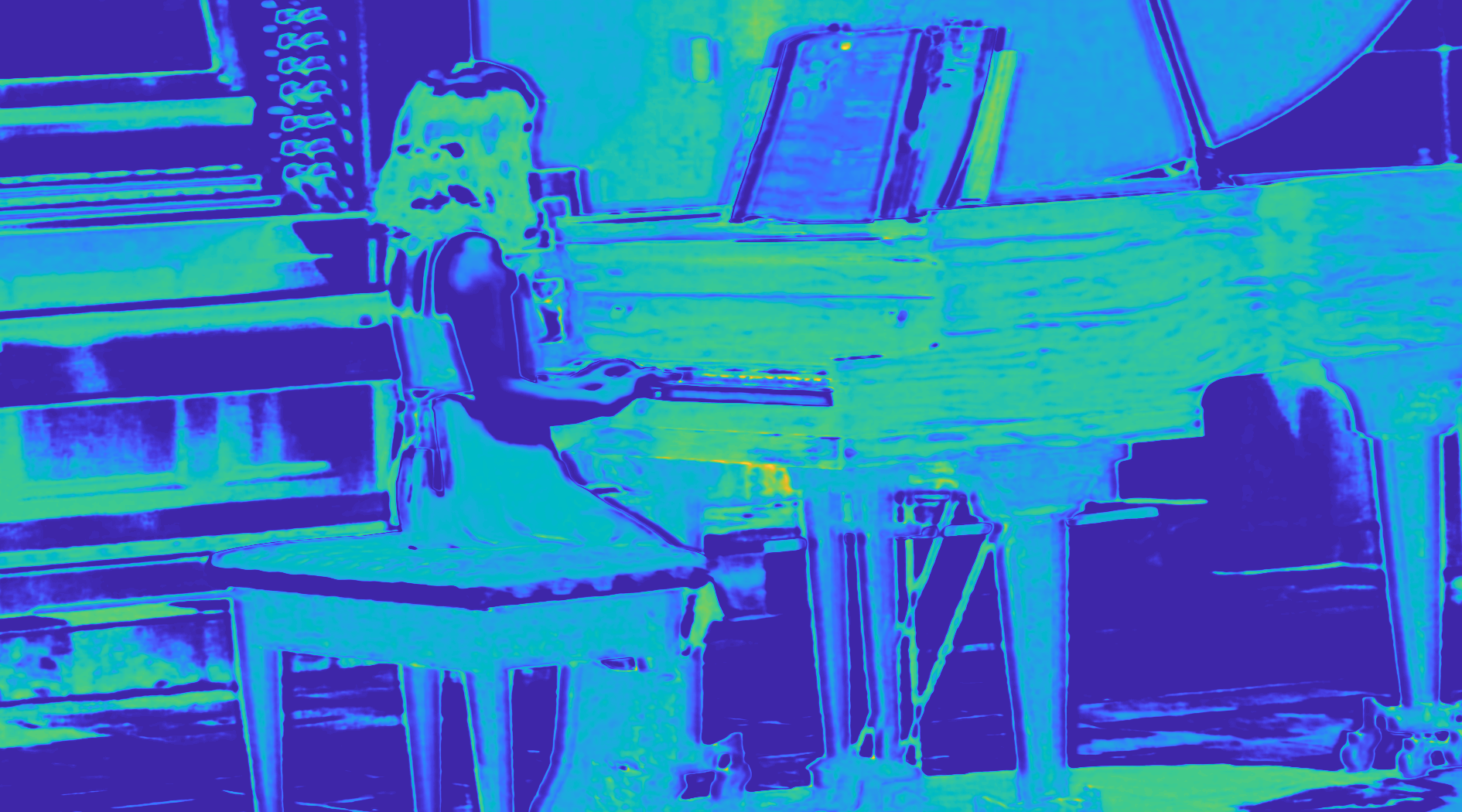} &
\includegraphics[width=0.23\textwidth]{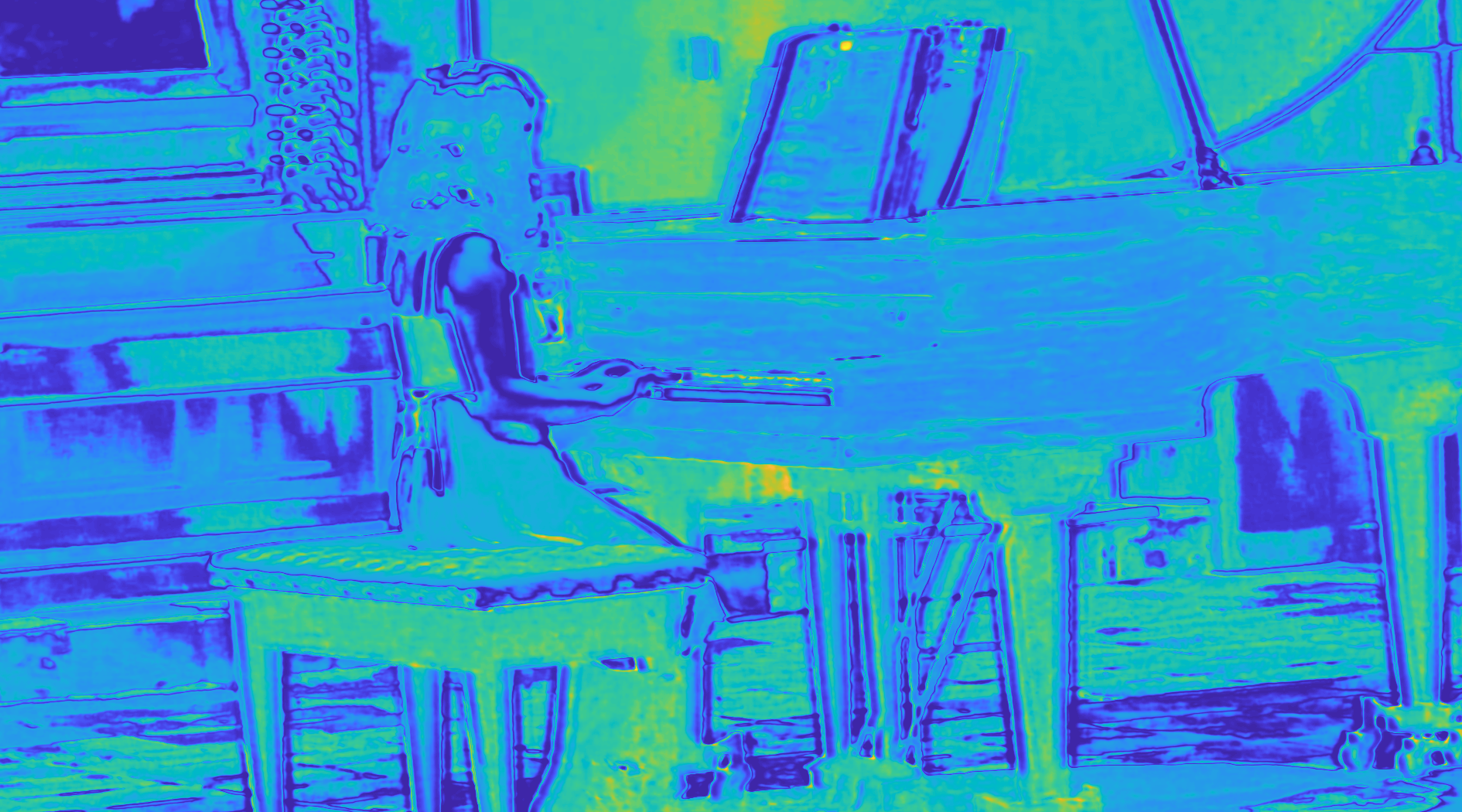} &
\includegraphics[width=0.23\textwidth]{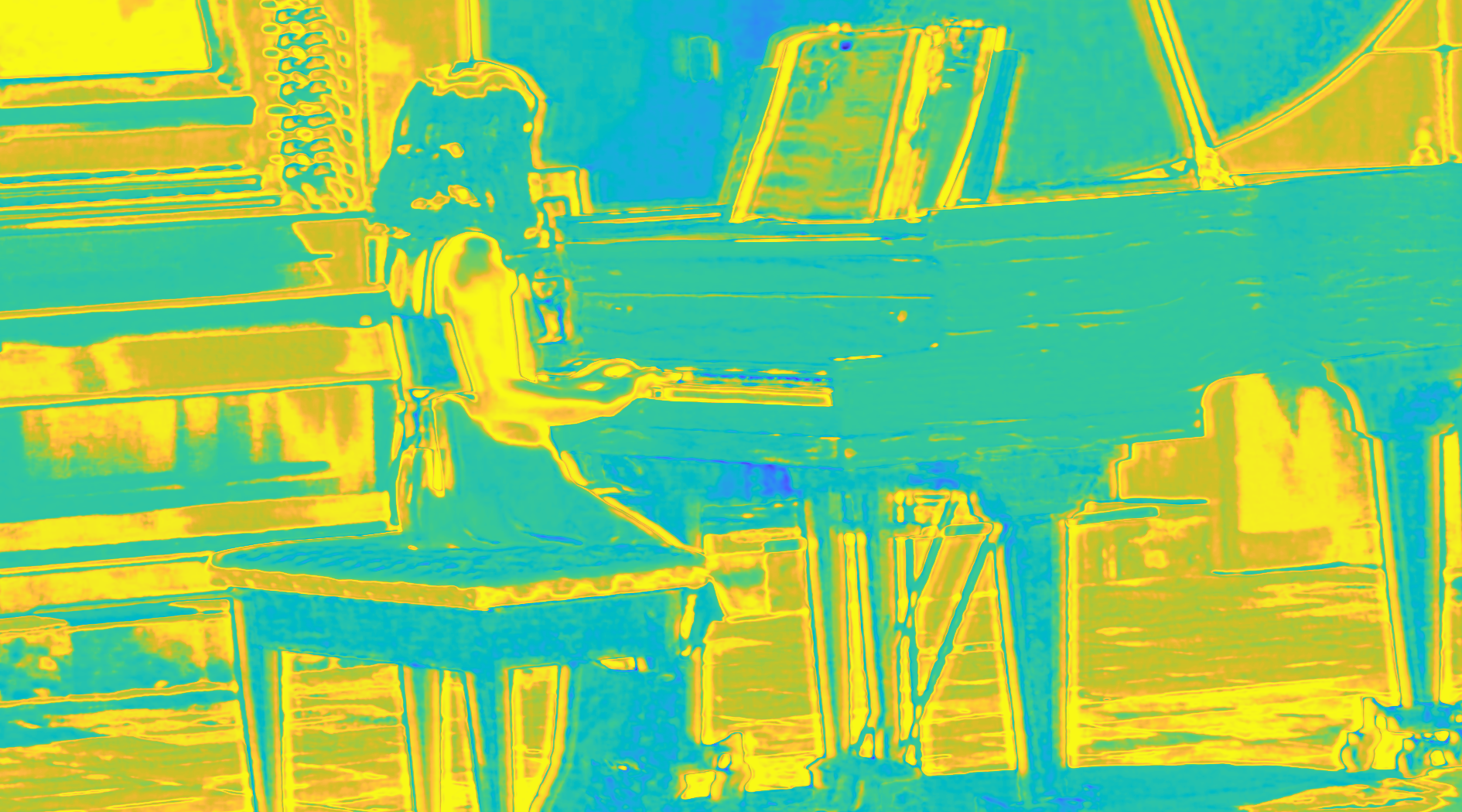} & \\ [-3pt]
& (3)  & (4)  & (5) \\
\end{tabular}
\begin{tabular}{@{}cccc@{}}
\includegraphics[width=0.23\textwidth]{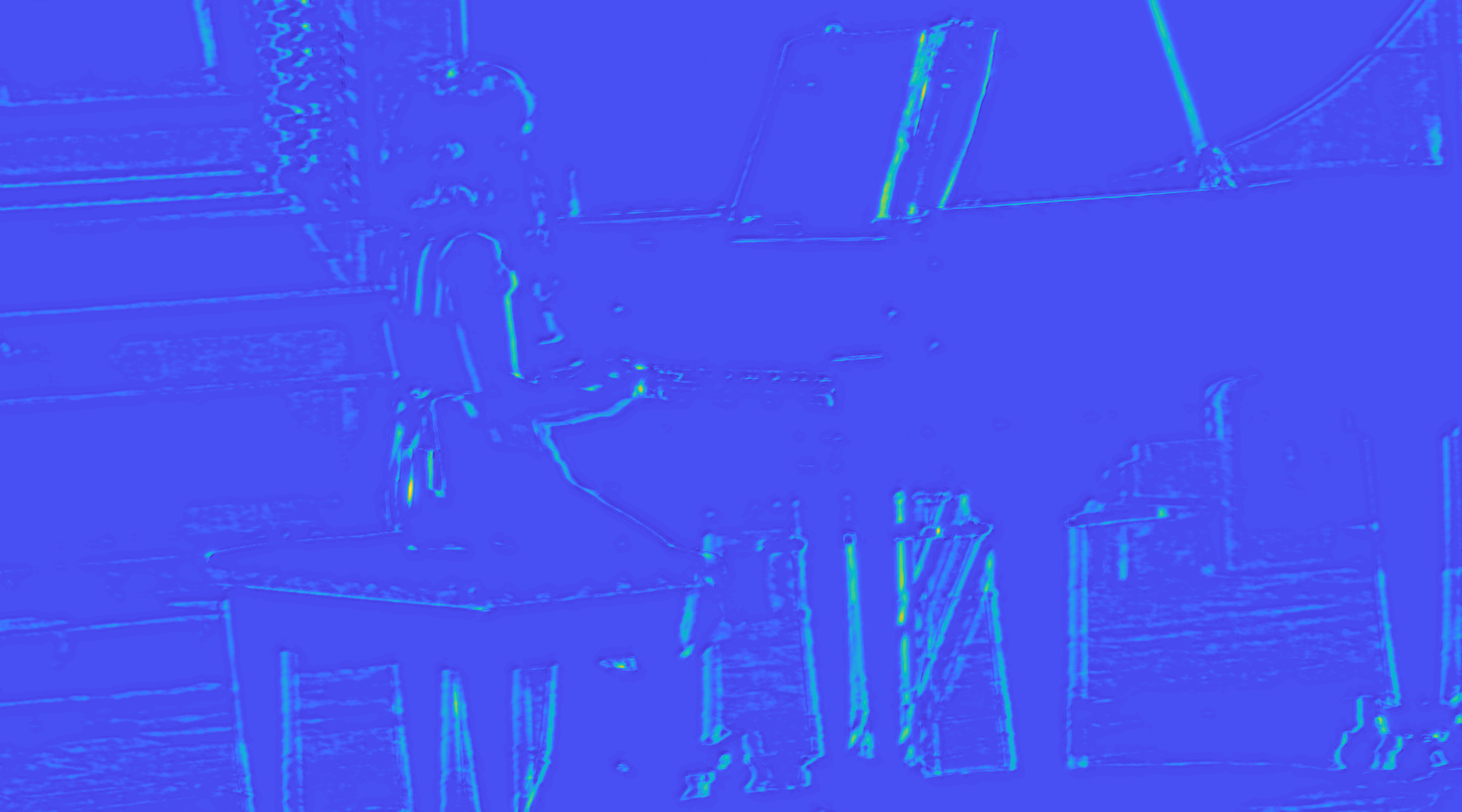} &
\includegraphics[width=0.23\textwidth]{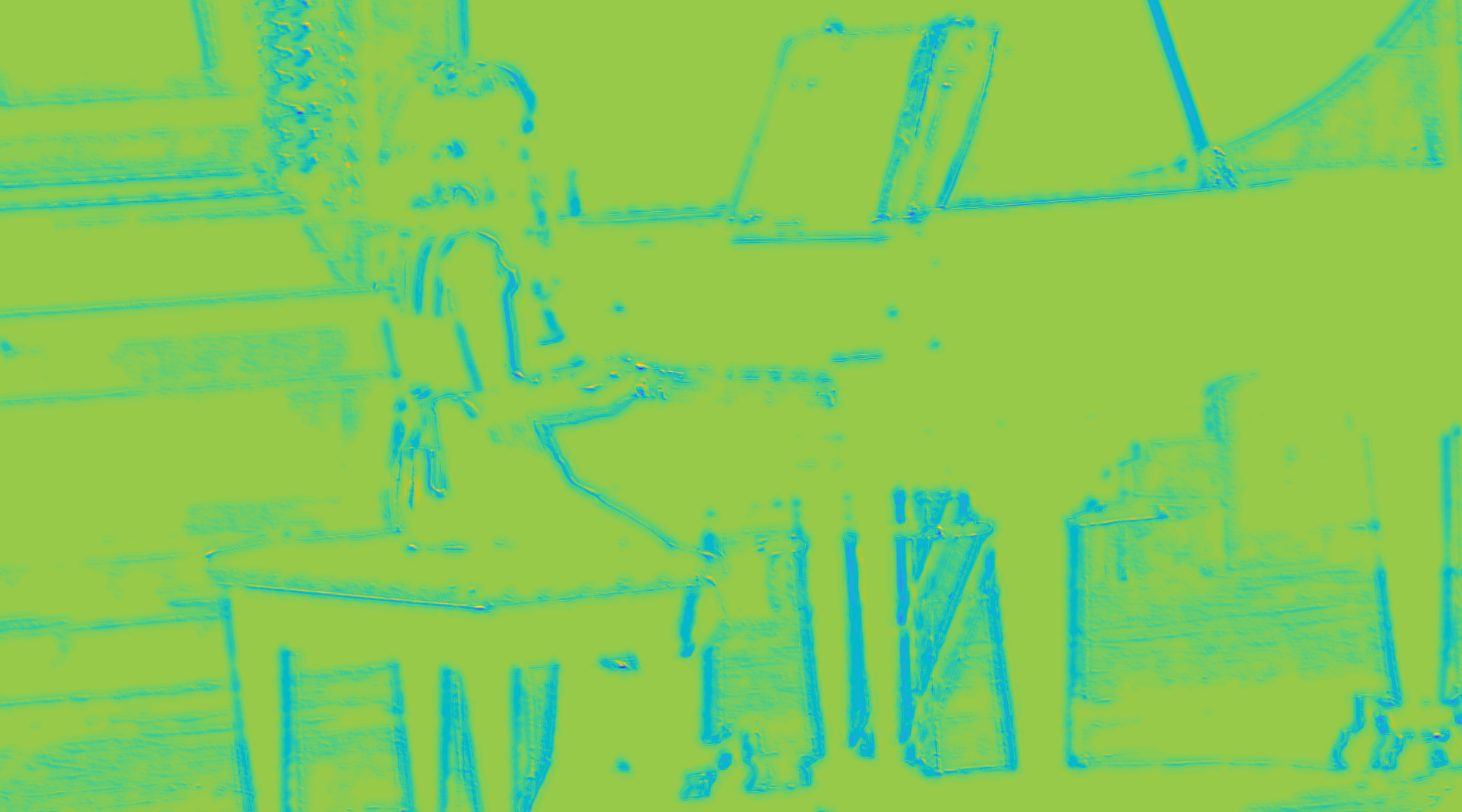} &
\includegraphics[width=0.23\textwidth]{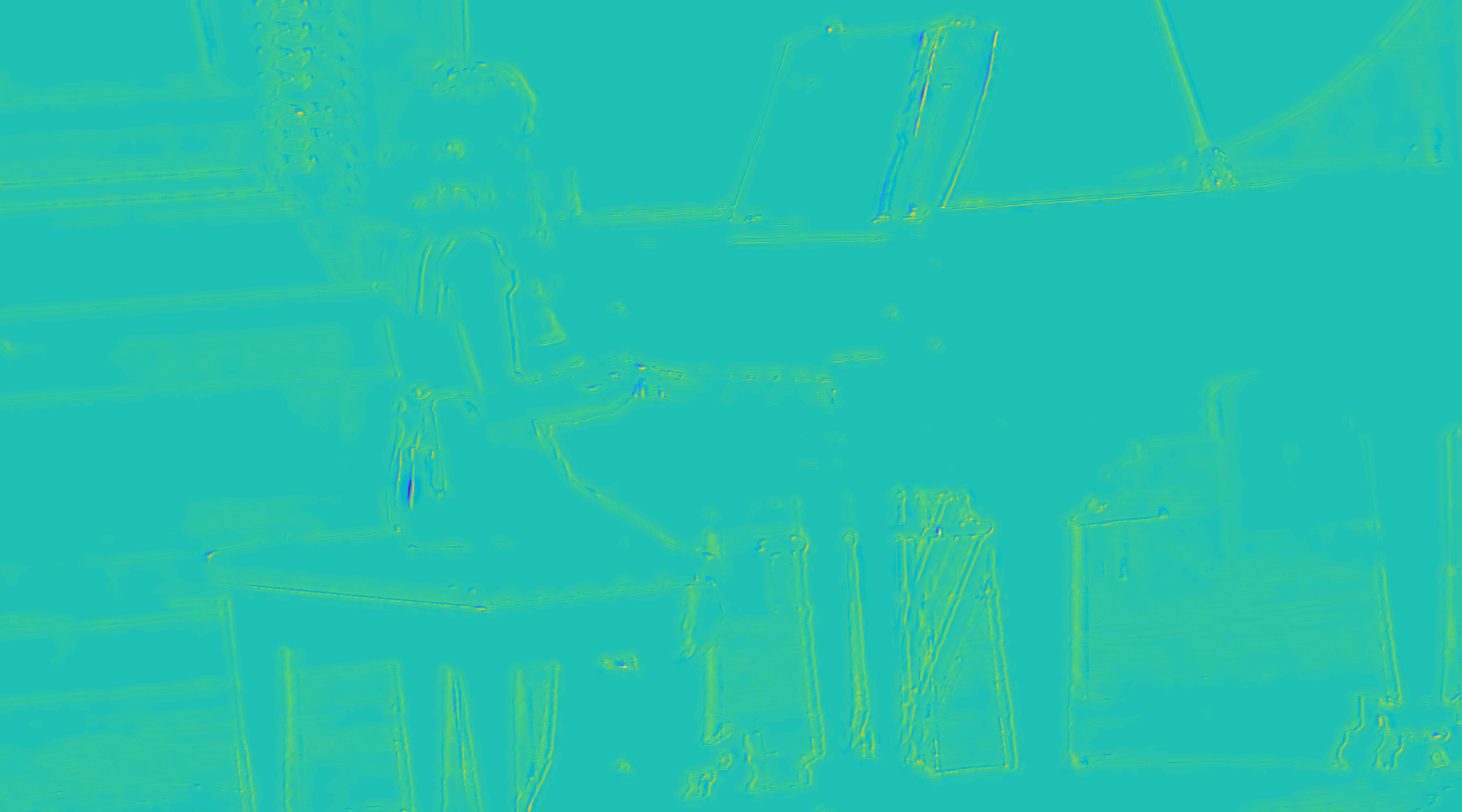} &
\includegraphics[width=0.23\textwidth]{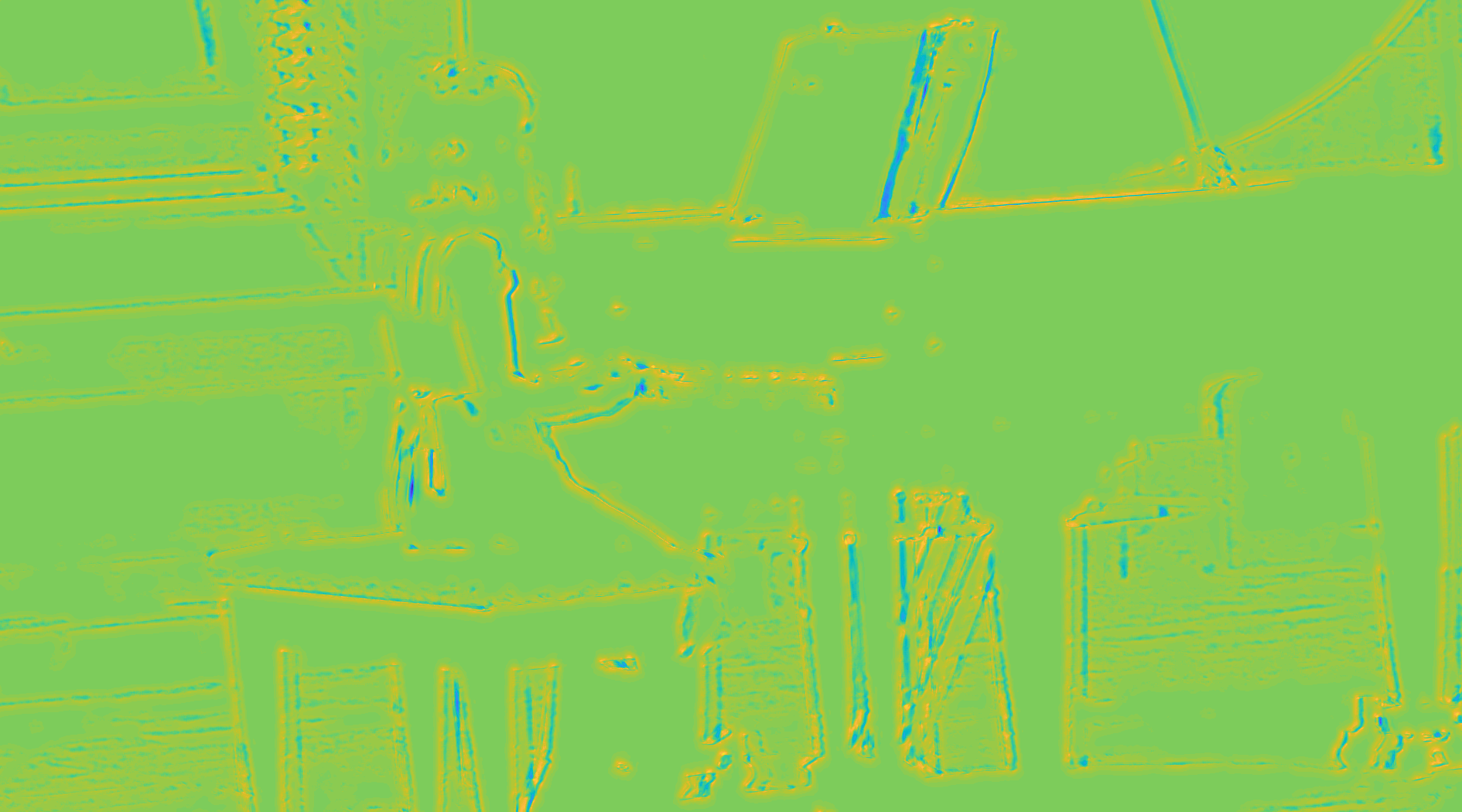} \\ [-3pt]
(6)  & (7)  & (8) & (9) \\
\end{tabular}
\begin{tabular}{@{}cccc@{}}
\includegraphics[width=0.23\textwidth]{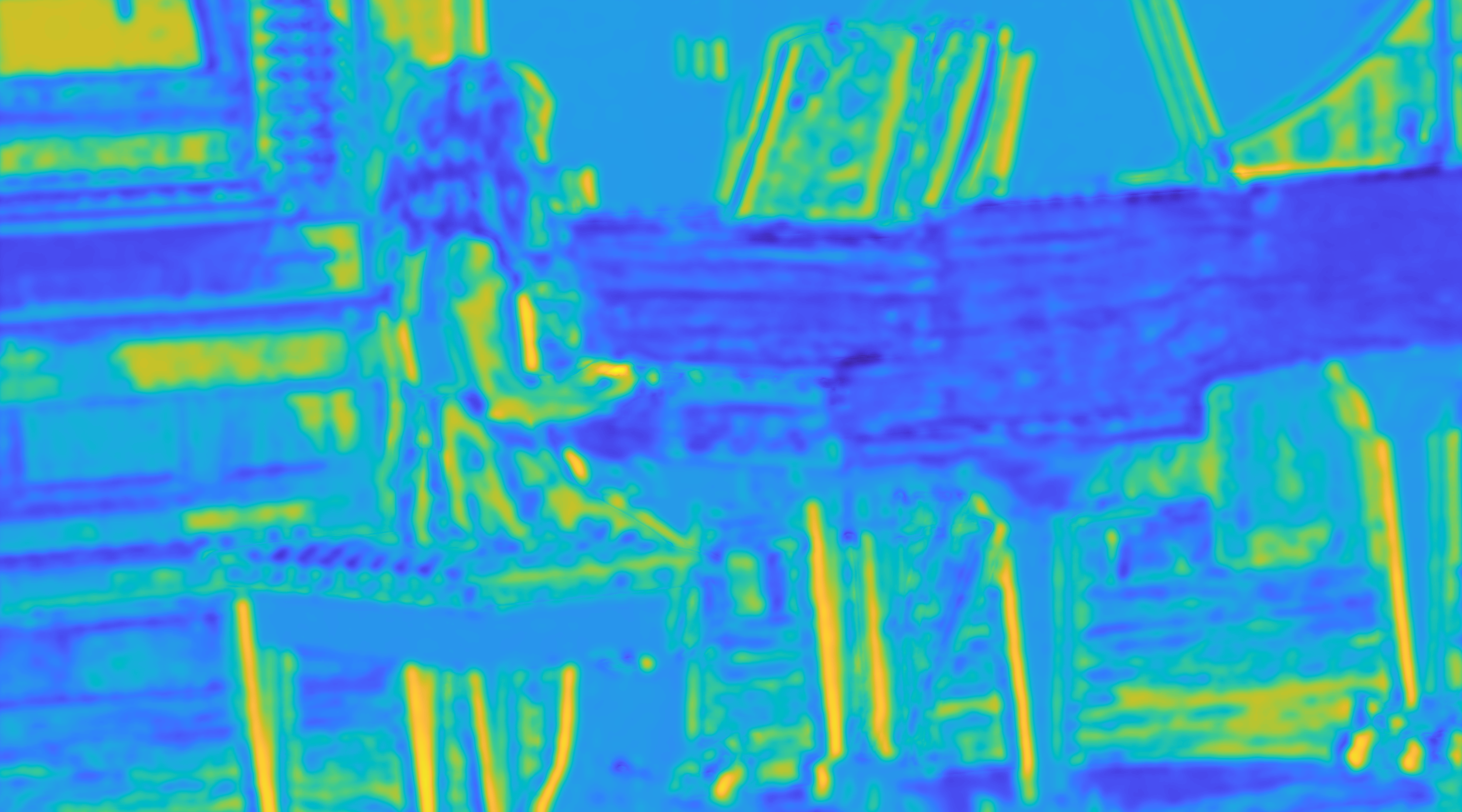} &
\includegraphics[width=0.23\textwidth]{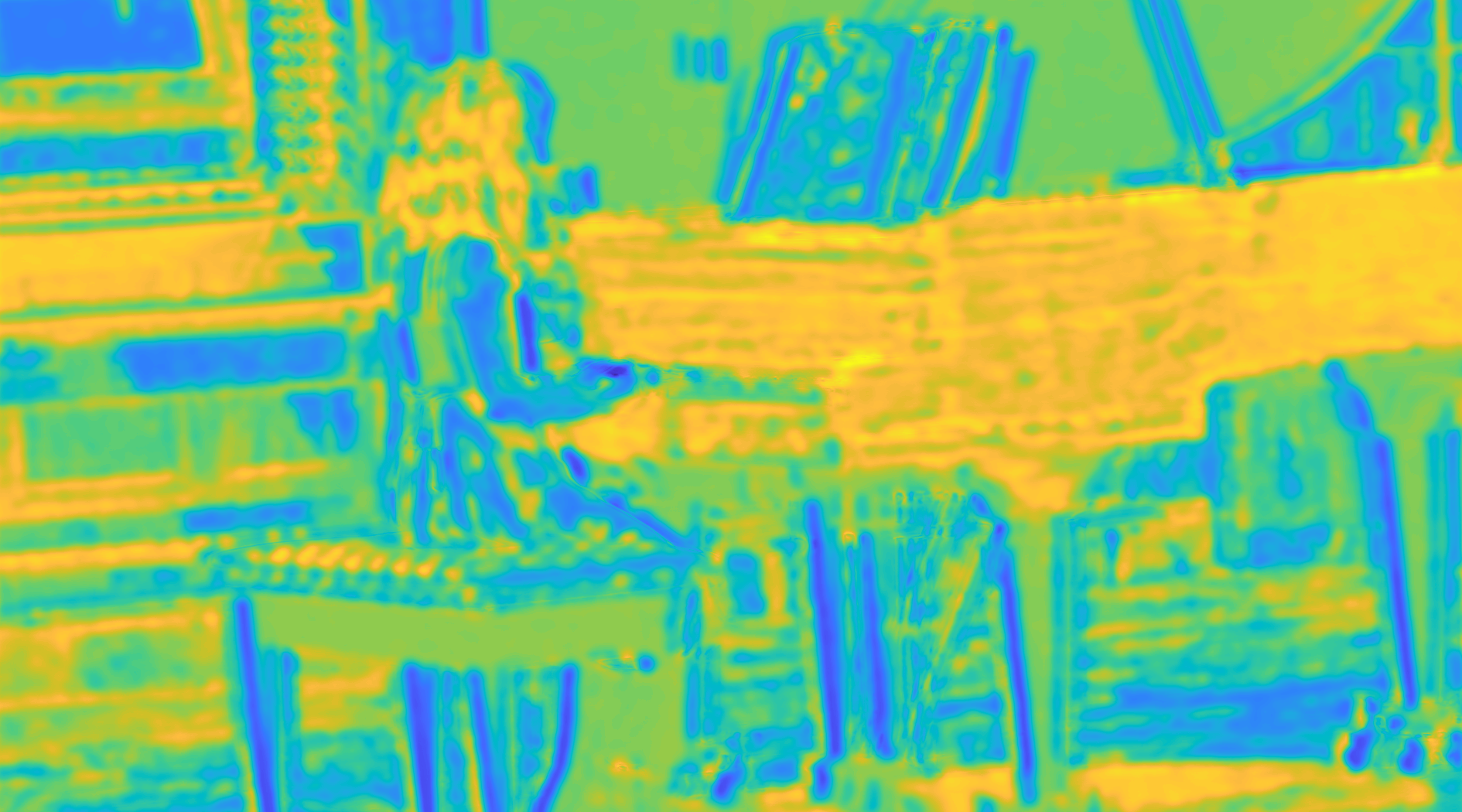} &
\includegraphics[width=0.23\textwidth]{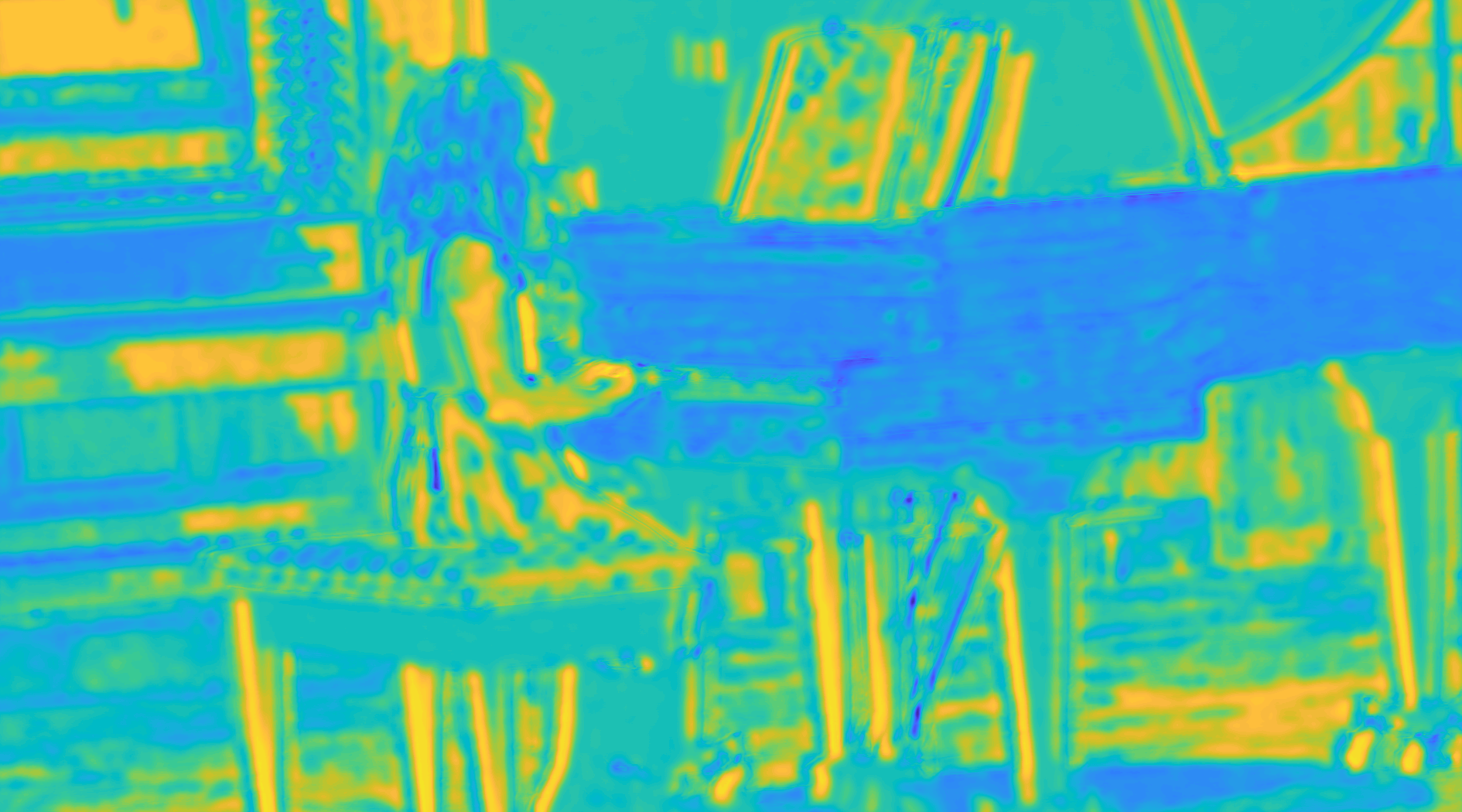} &
\includegraphics[width=0.23\textwidth]{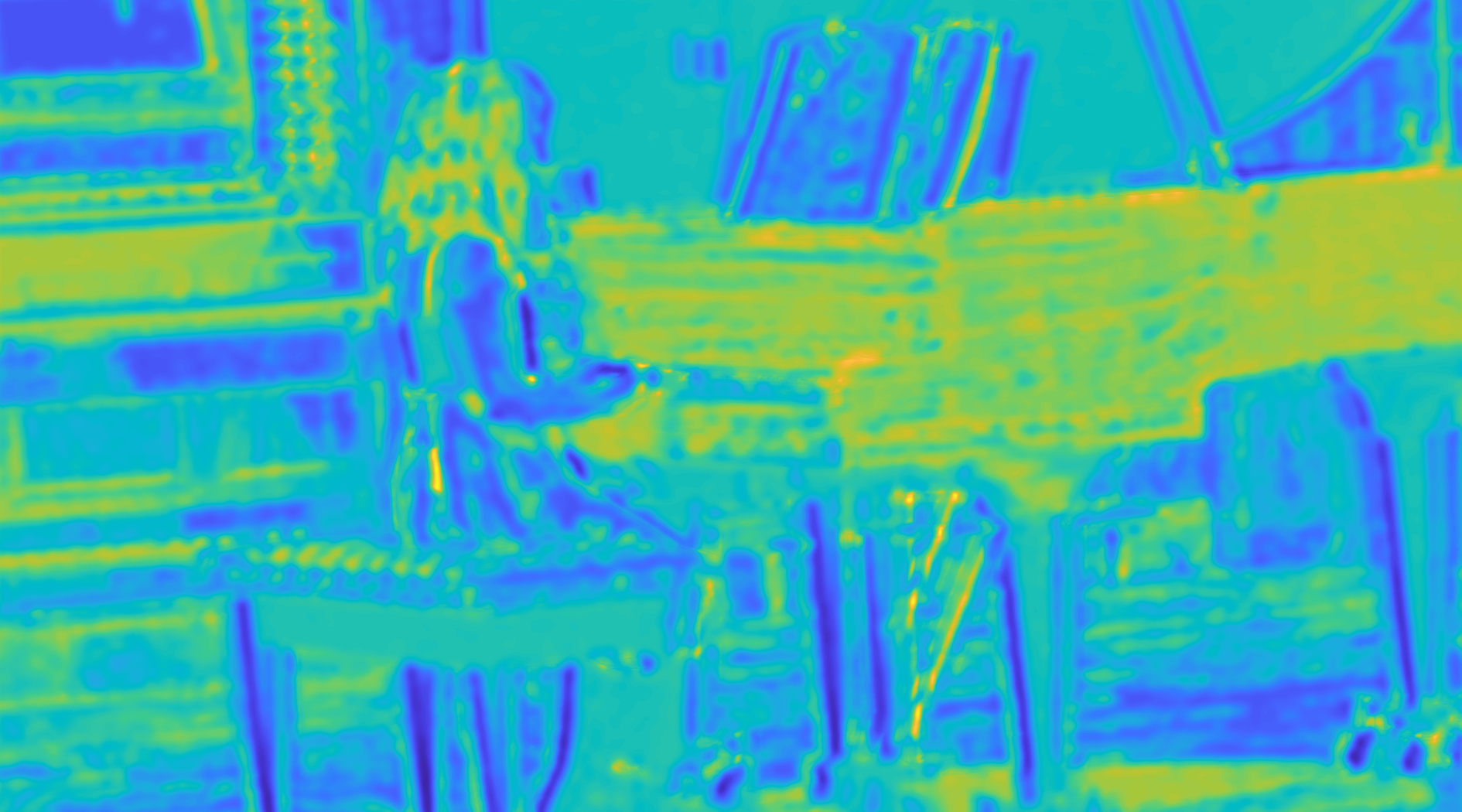} \\ [-3pt]
(10)  & (11)  & (12) & (13) \\
\end{tabular}
\begin{tabular}{@{}cccc@{}}
\includegraphics[width=0.23\textwidth]{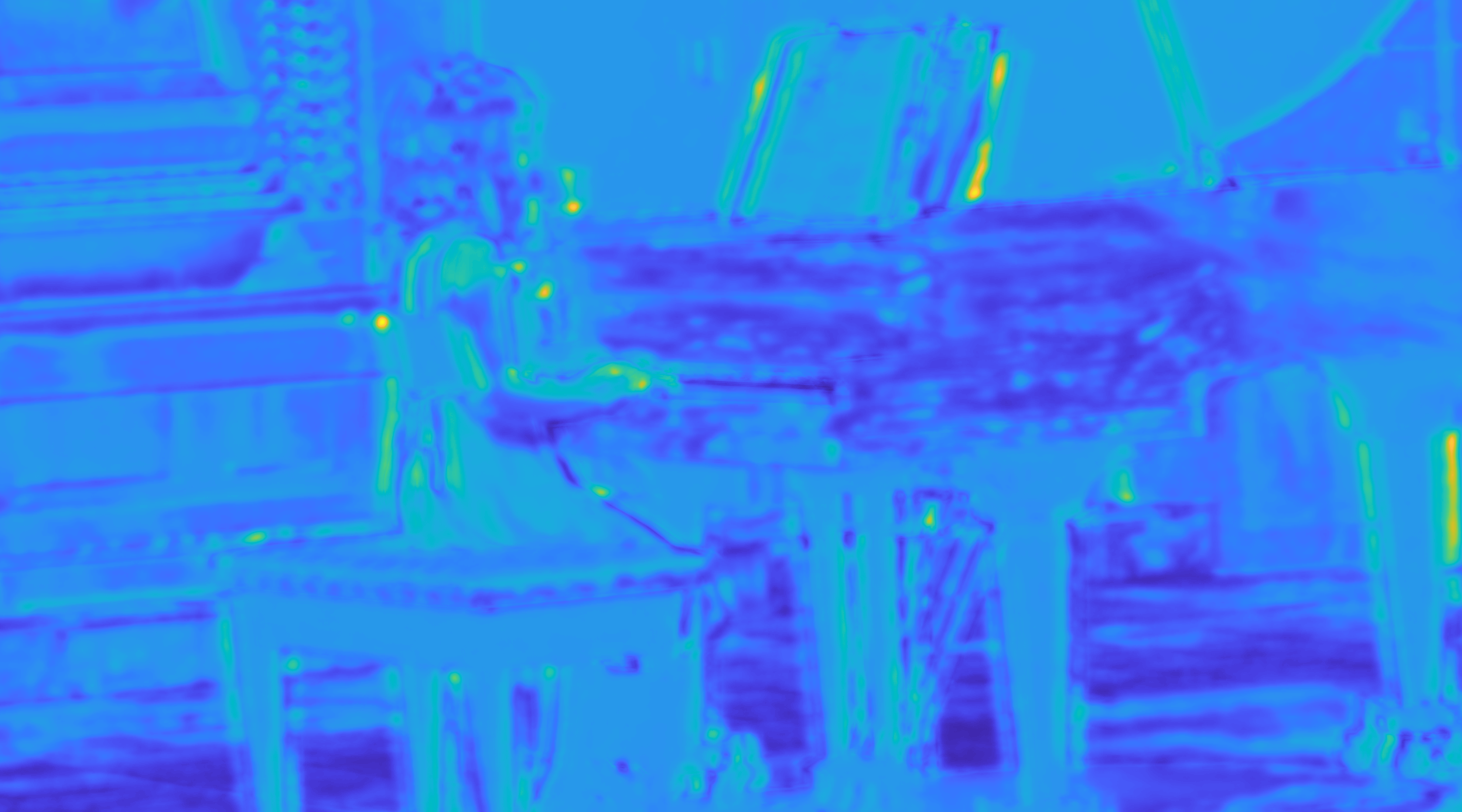} &
\includegraphics[width=0.23\textwidth]{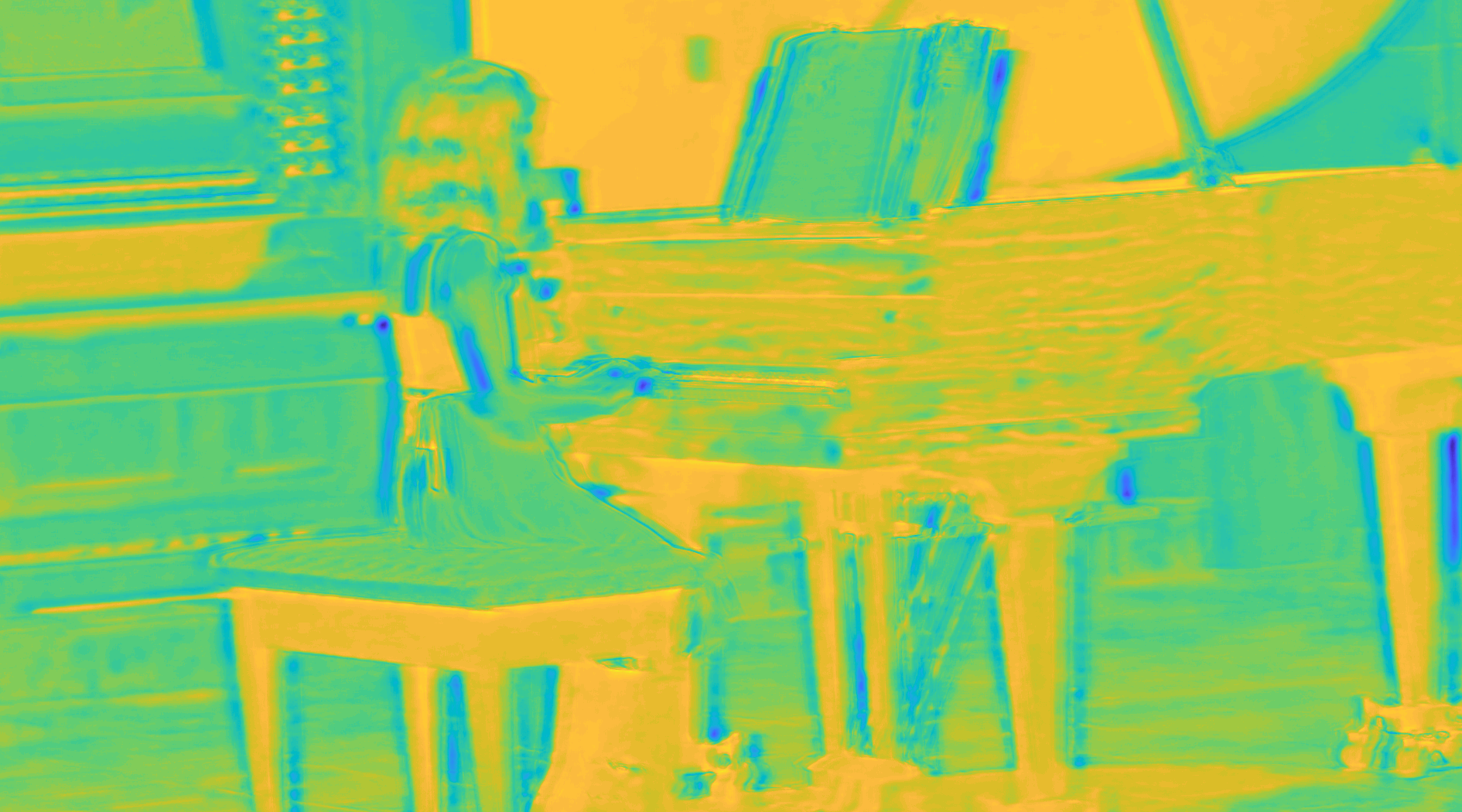} &
\includegraphics[width=0.23\textwidth]{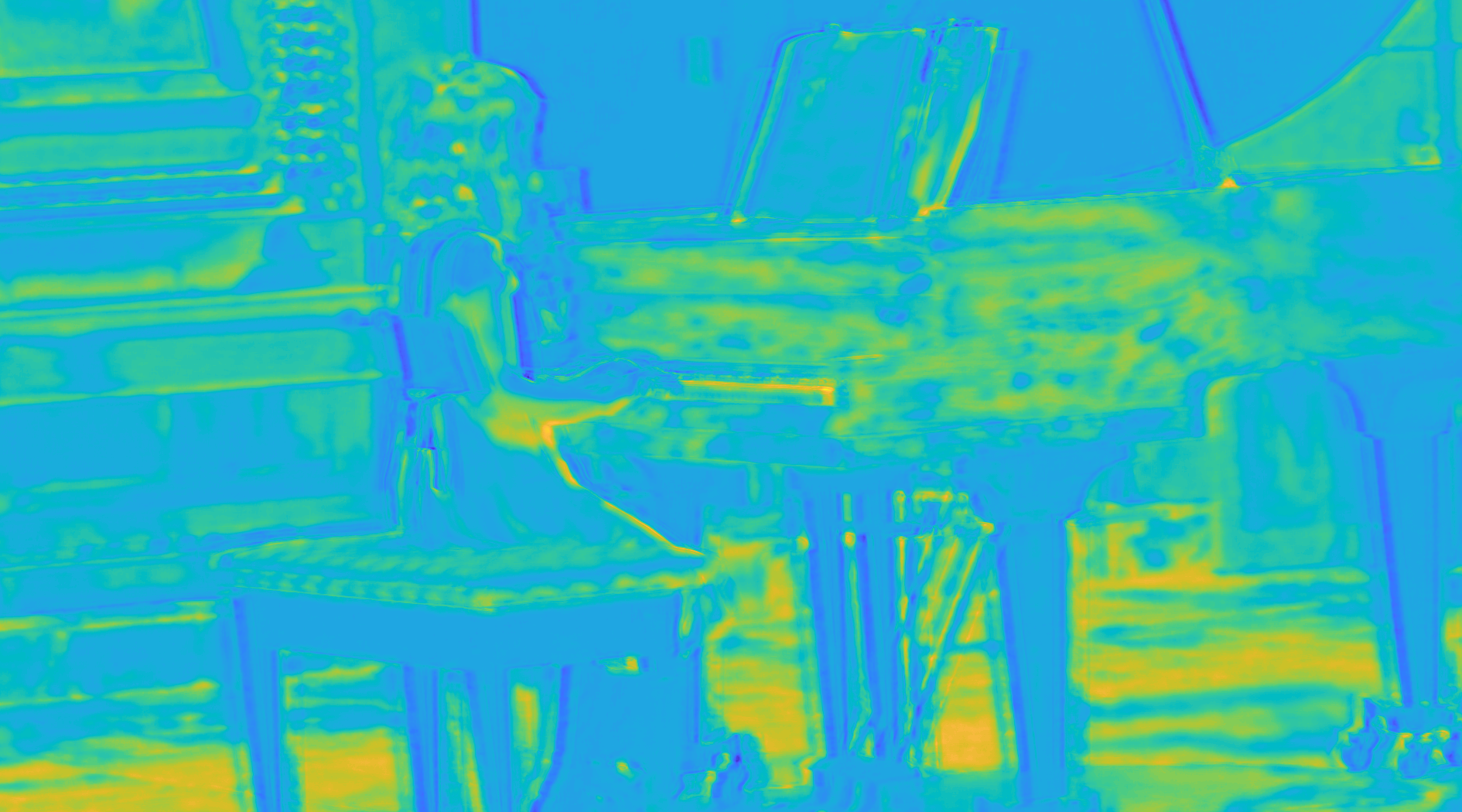} &
\includegraphics[width=0.23\textwidth]{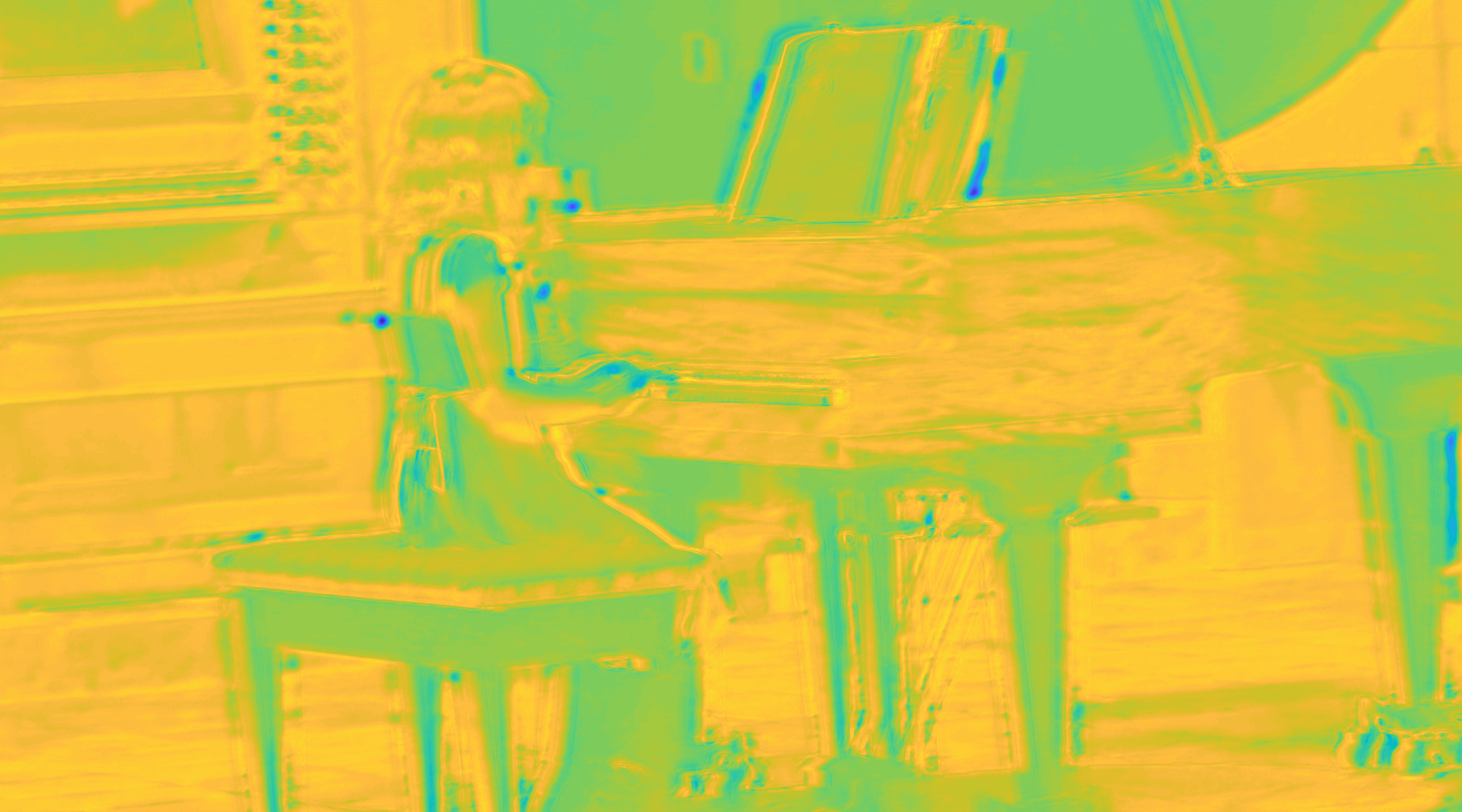} \\ [-3pt]
(14)  & (15)  & (16) & (17) \\
\end{tabular}
\caption{Visualization of external and internal attention maps on real-world blur dataset proposed in \cite{su2017deep}. (1): blurry frame; (2): Deblurred frame; (3)-(5): attention maps of external attention branch 1-3; (6)-(9): attention maps of internal attention branch 1-3 in the first internal attention module; (10)-(13): attention maps of internal attention branch 1-3 in the second internal attention module; (14)-(17): attention maps of internal attention branch 1-3 in the third internal attention module.}
\label{fig:attention_map}
\end{figure*}

\begin{figure*}[htp!]
\vspace{-1em}
\centering
\setlength\tabcolsep{1.5pt}
\begin{tabular}{@{}cccccc@{}}
\includegraphics[width=0.16\textwidth]{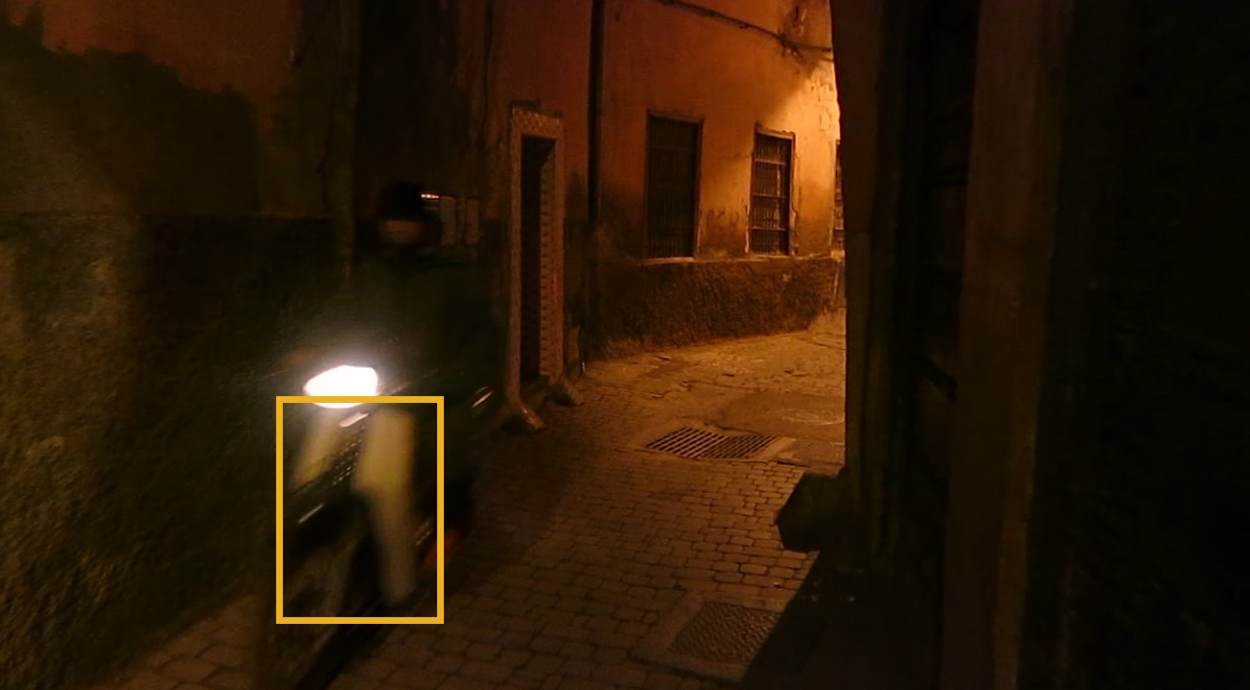} &
\includegraphics[width=0.16\textwidth]{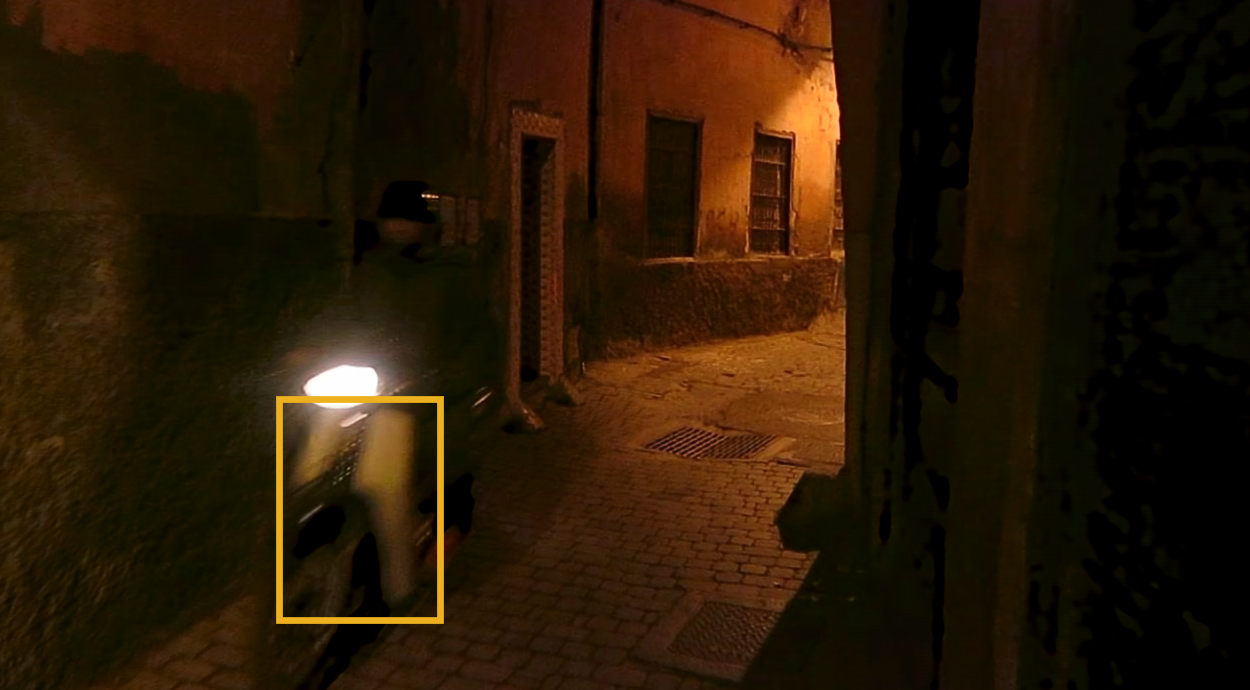} &
\includegraphics[width=0.16\textwidth]{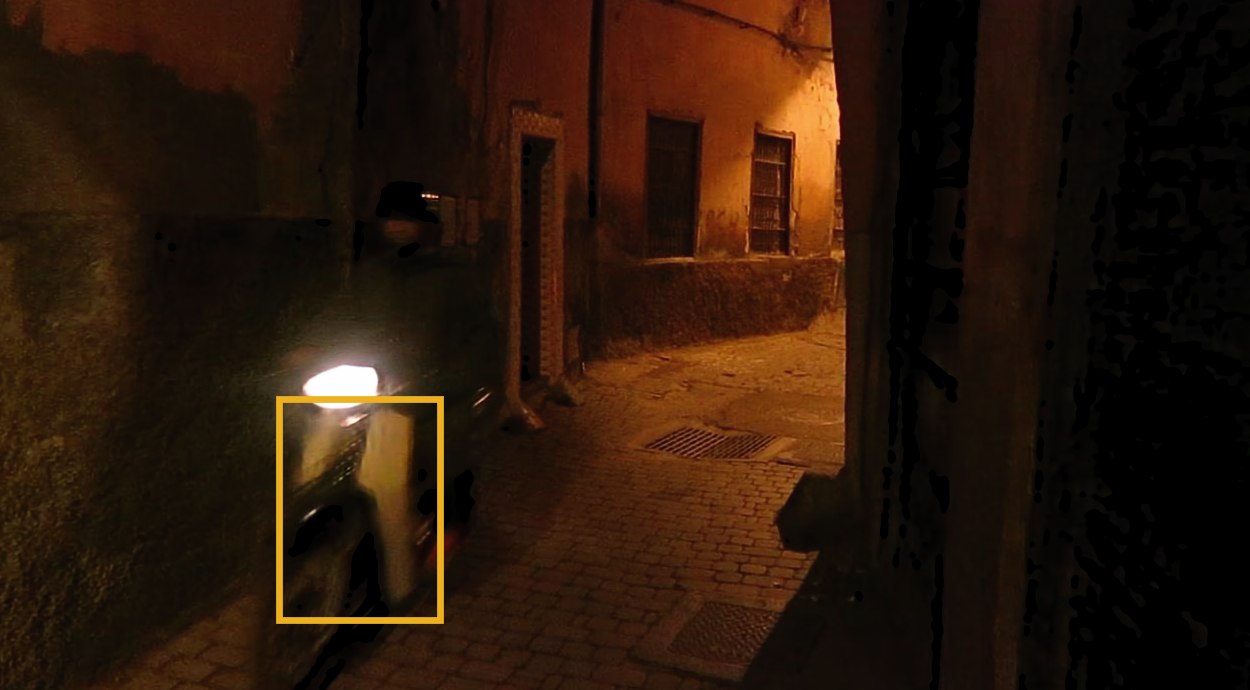} &
\includegraphics[width=0.16\textwidth]{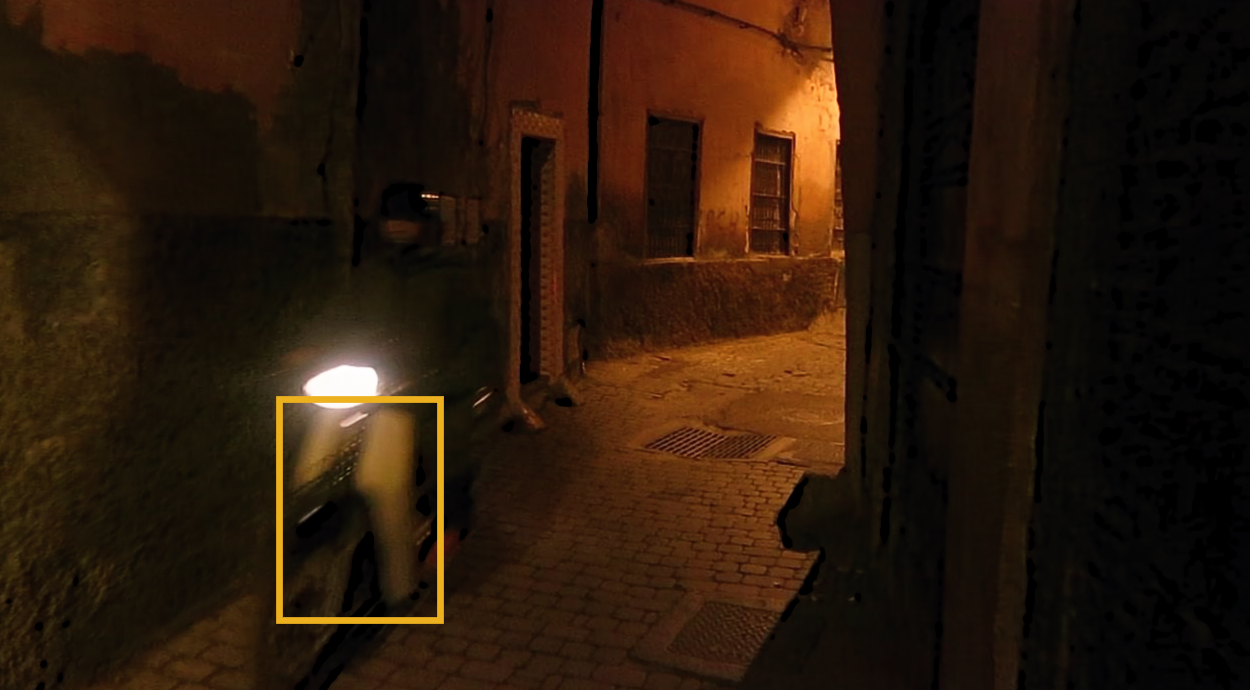} &
\includegraphics[width=0.16\textwidth]{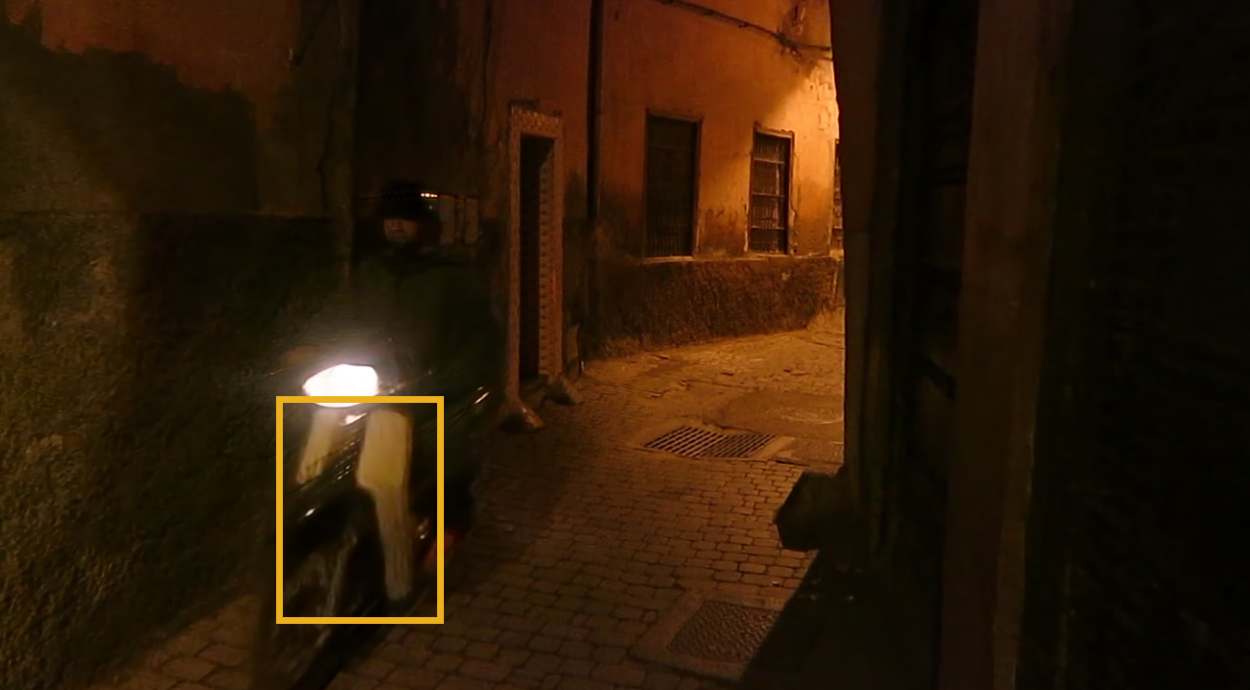} &
\includegraphics[width=0.16\textwidth]{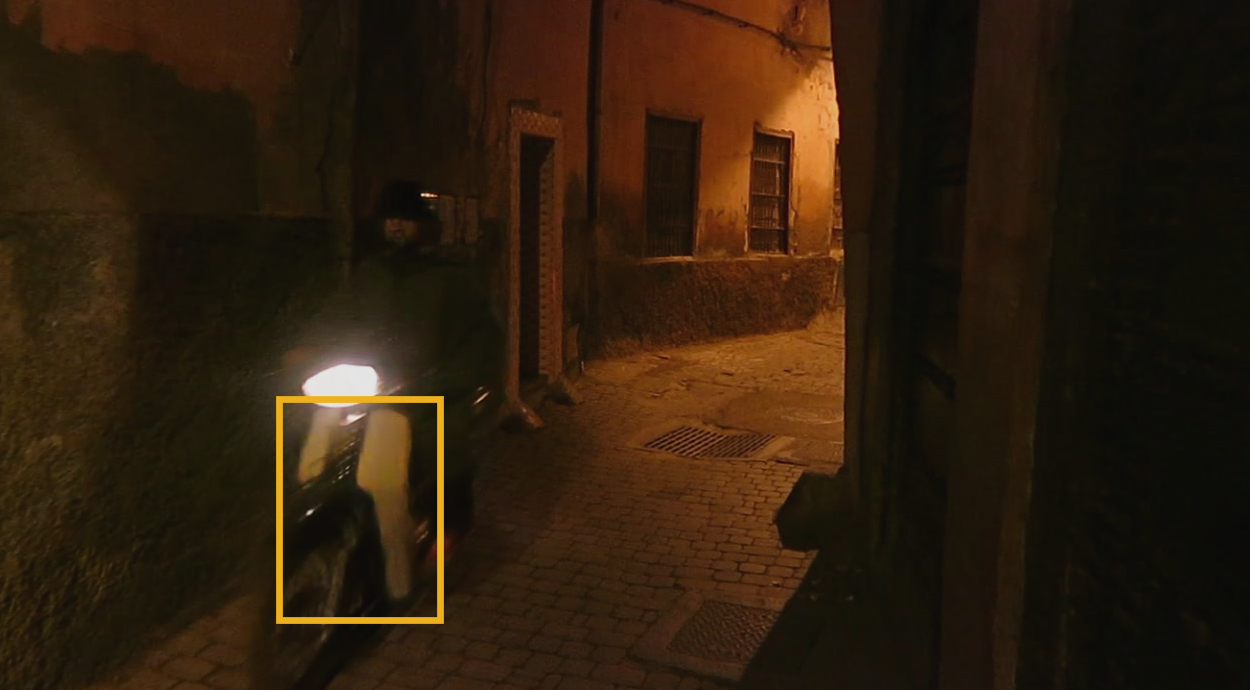} \\ [-3pt]
\end{tabular}
\begin{tabular}{@{}cccccc@{}}
\includegraphics[width=0.16\textwidth]{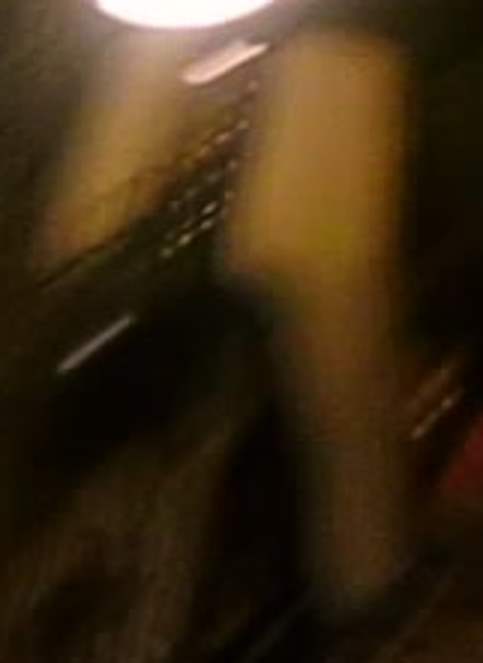} &
\includegraphics[width=0.16\textwidth]{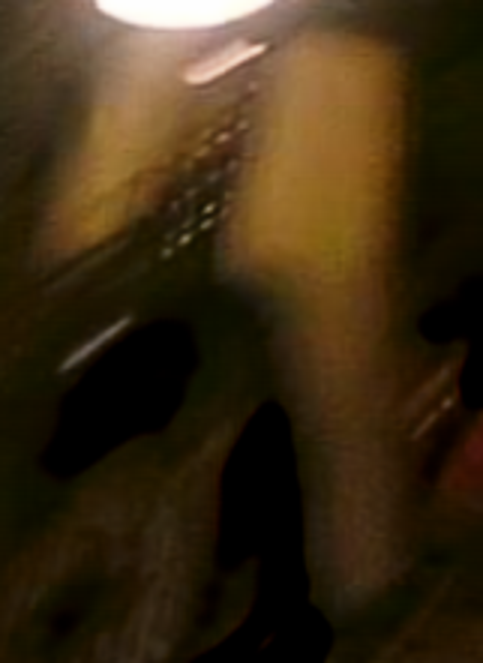} &
\includegraphics[width=0.16\textwidth]{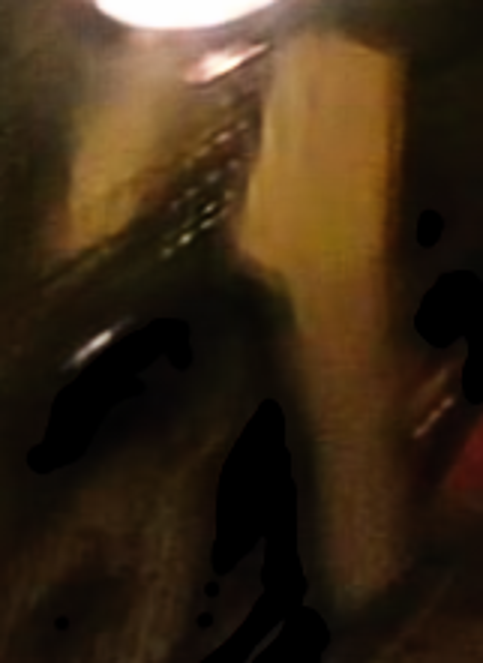} &
\includegraphics[width=0.16\textwidth]{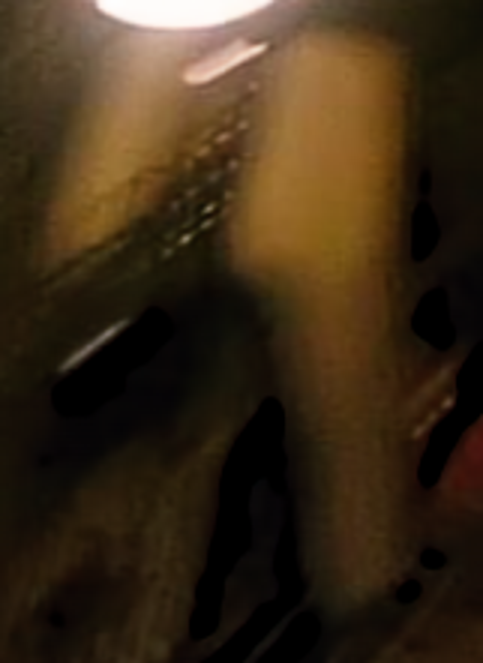} &
\includegraphics[width=0.16\textwidth]{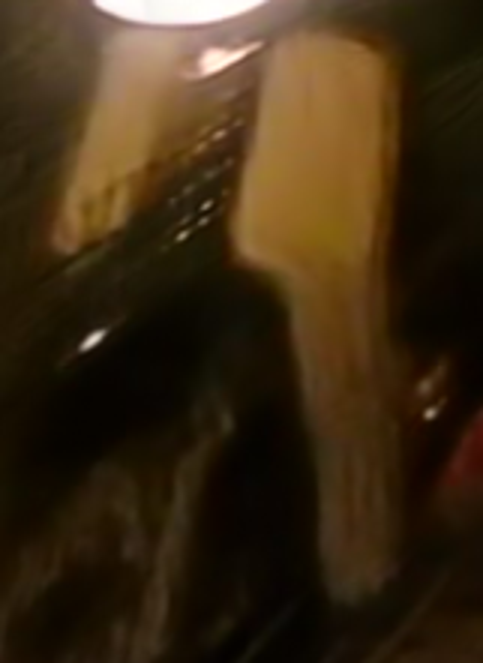} &
\includegraphics[width=0.16\textwidth]{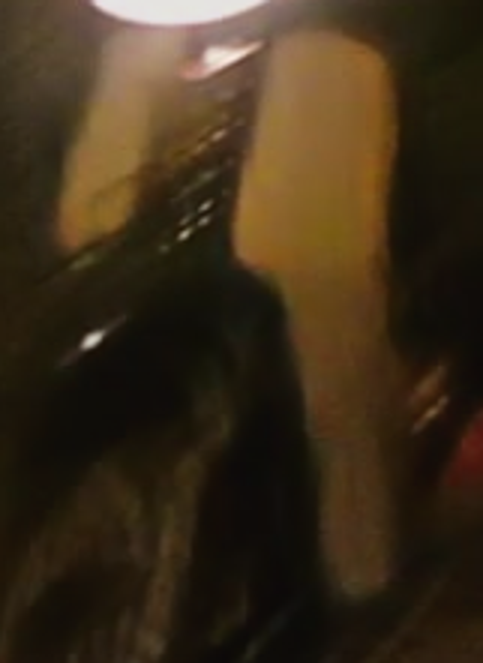} \\ [-3pt]
\end{tabular}
\begin{tabular}{@{}cccccc@{}}
\includegraphics[width=0.16\textwidth]{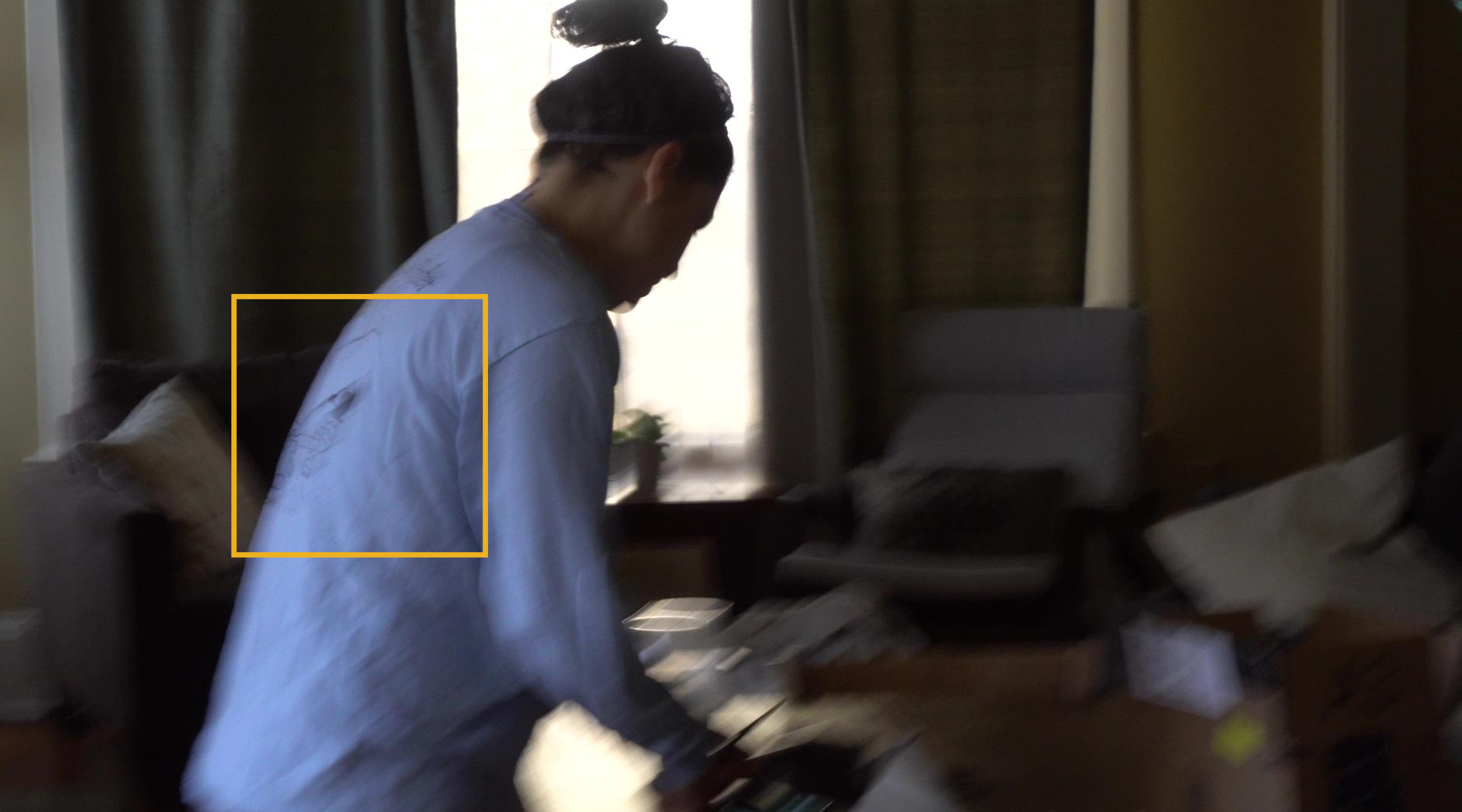} &
\includegraphics[width=0.16\textwidth]{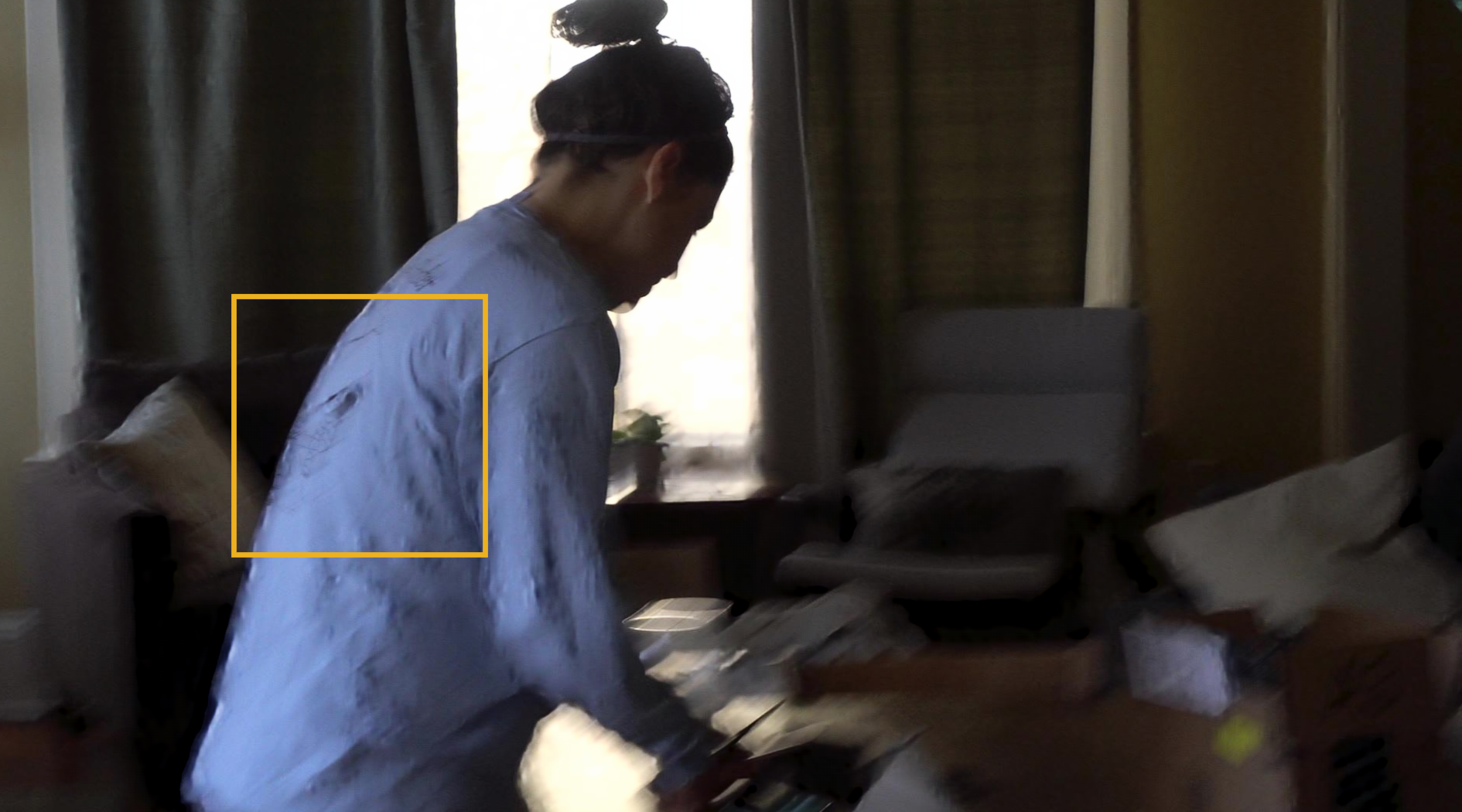} &
\includegraphics[width=0.16\textwidth]{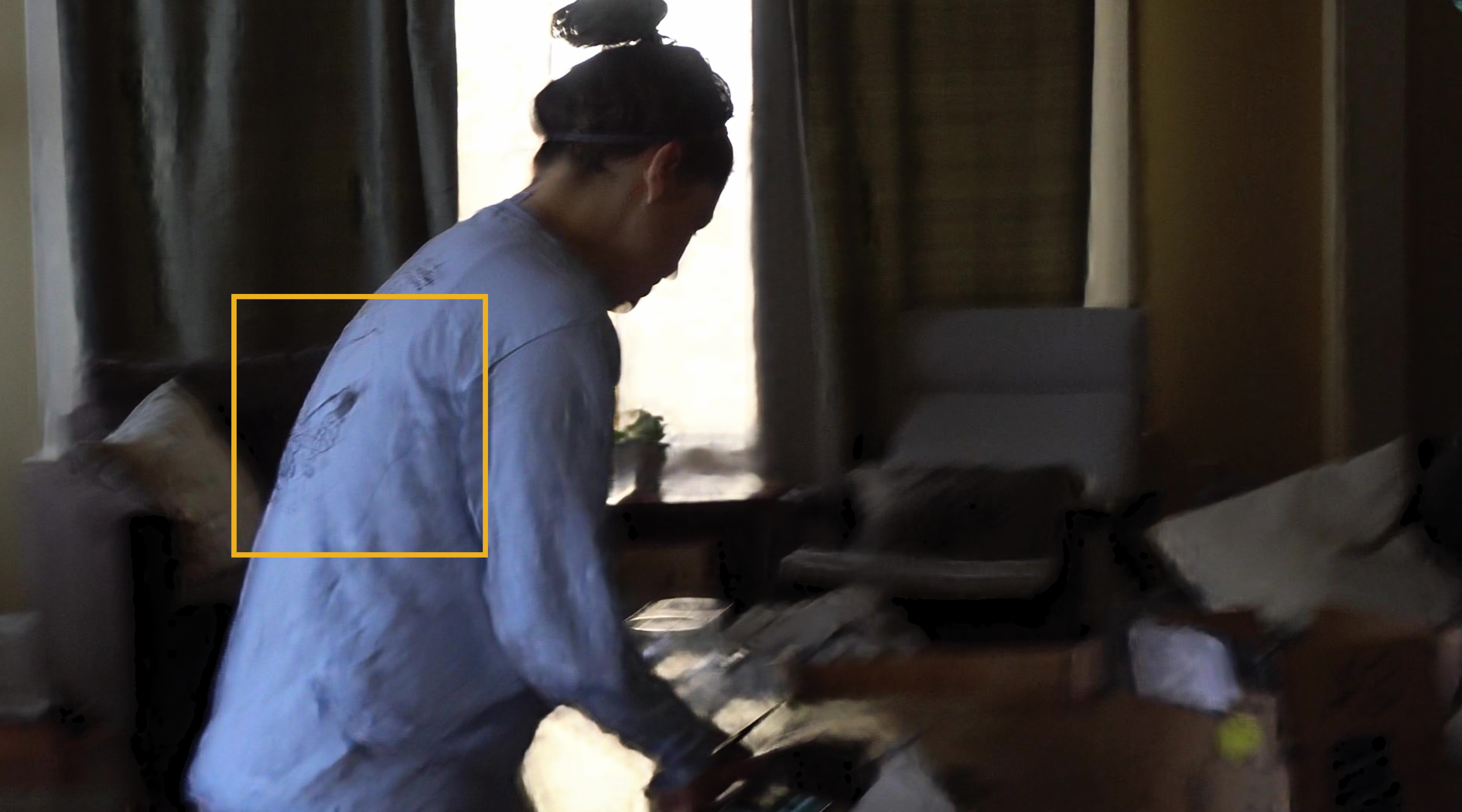} &
\includegraphics[width=0.16\textwidth]{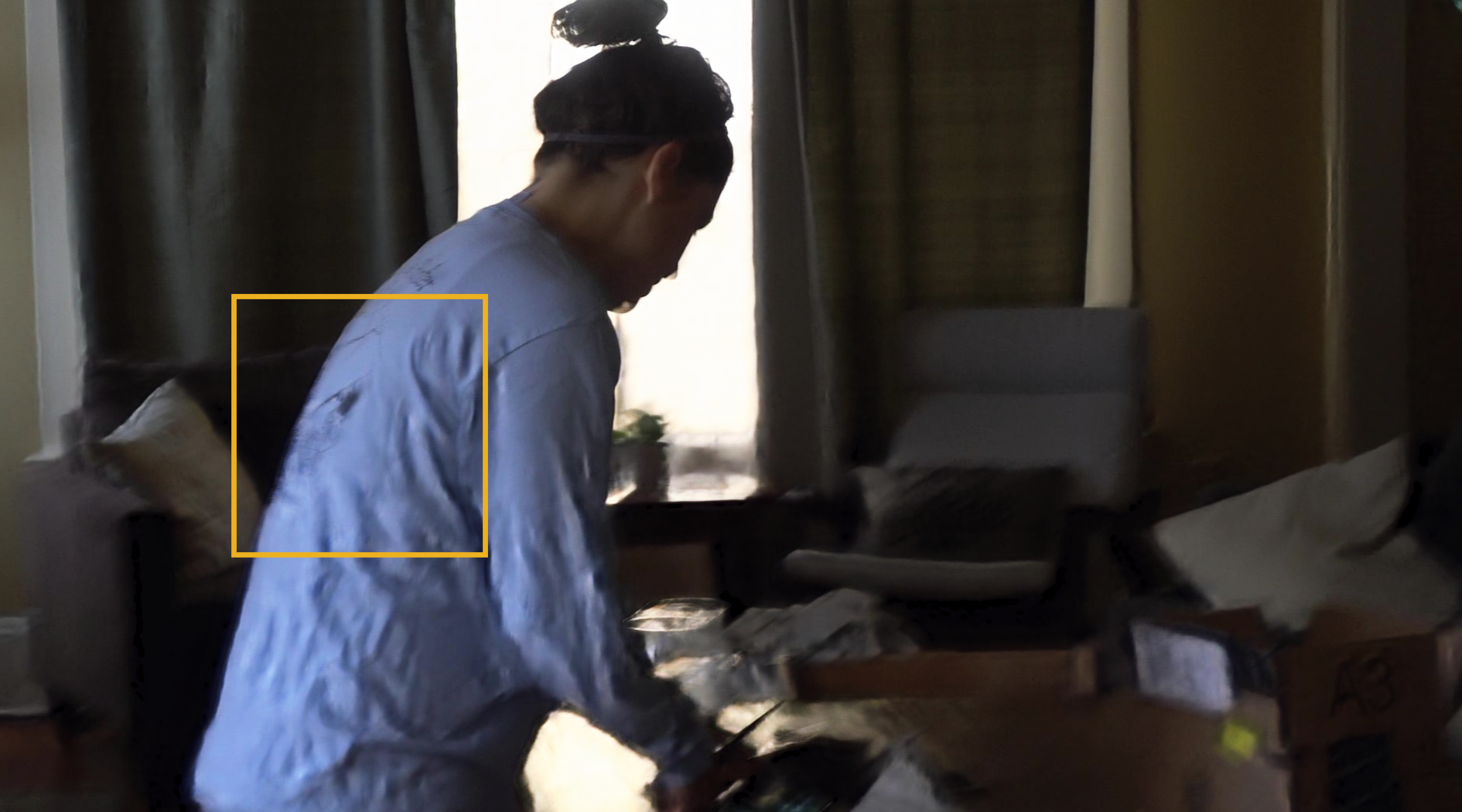} &
\includegraphics[width=0.16\textwidth]{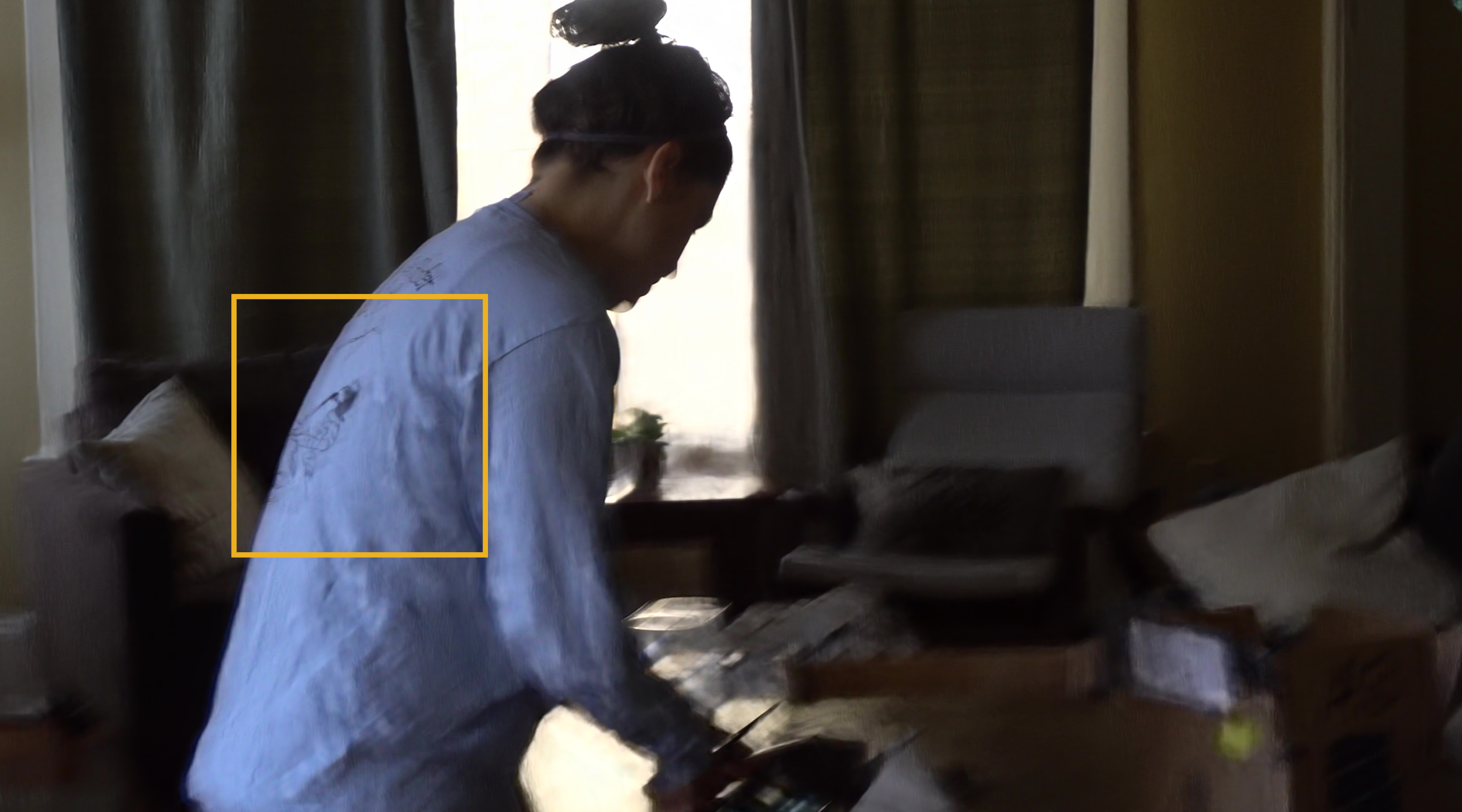} &
\includegraphics[width=0.16\textwidth]{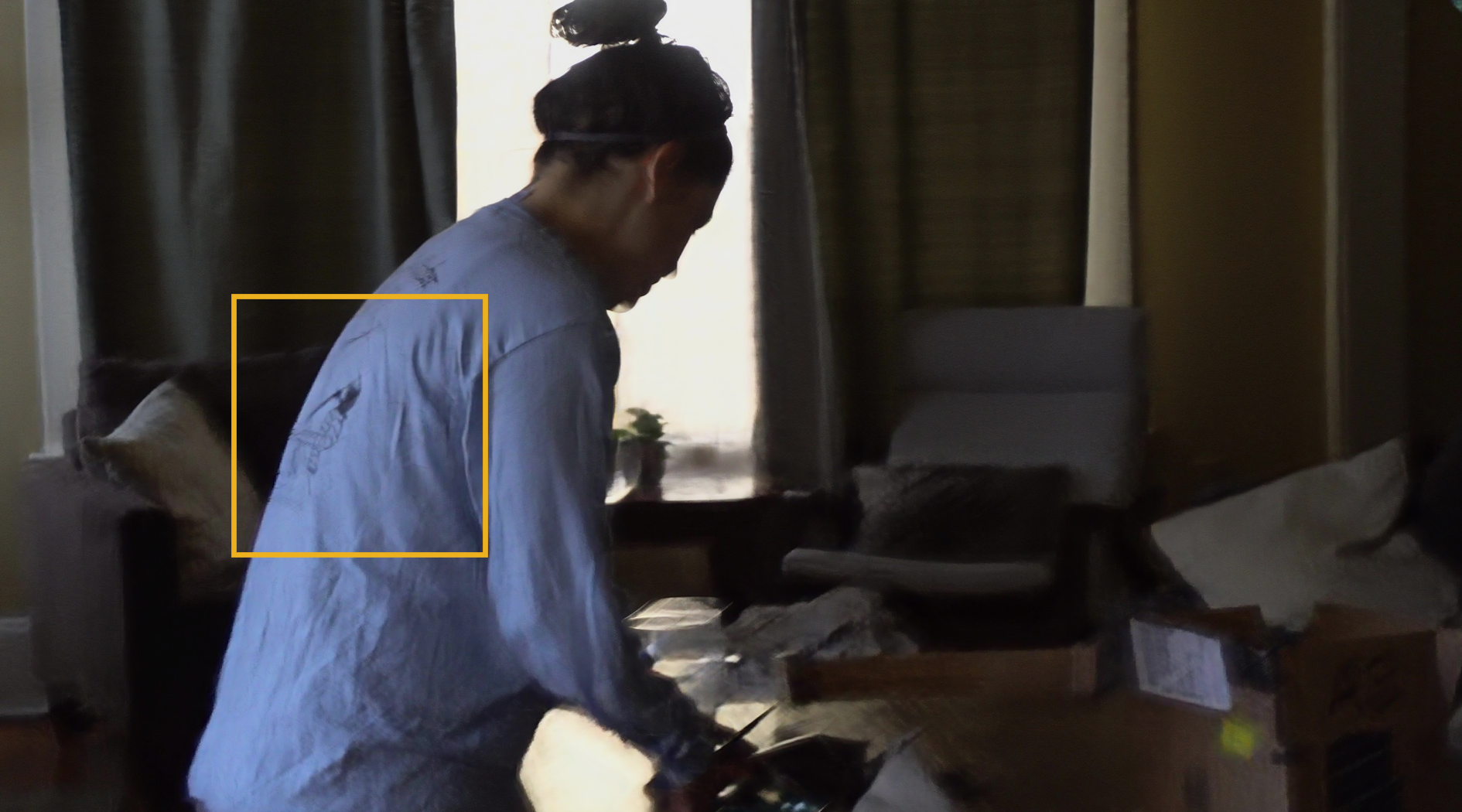} \\ [-3pt]
\end{tabular}
\begin{tabular}{@{}cccccc@{}}
\includegraphics[width=0.16\textwidth]{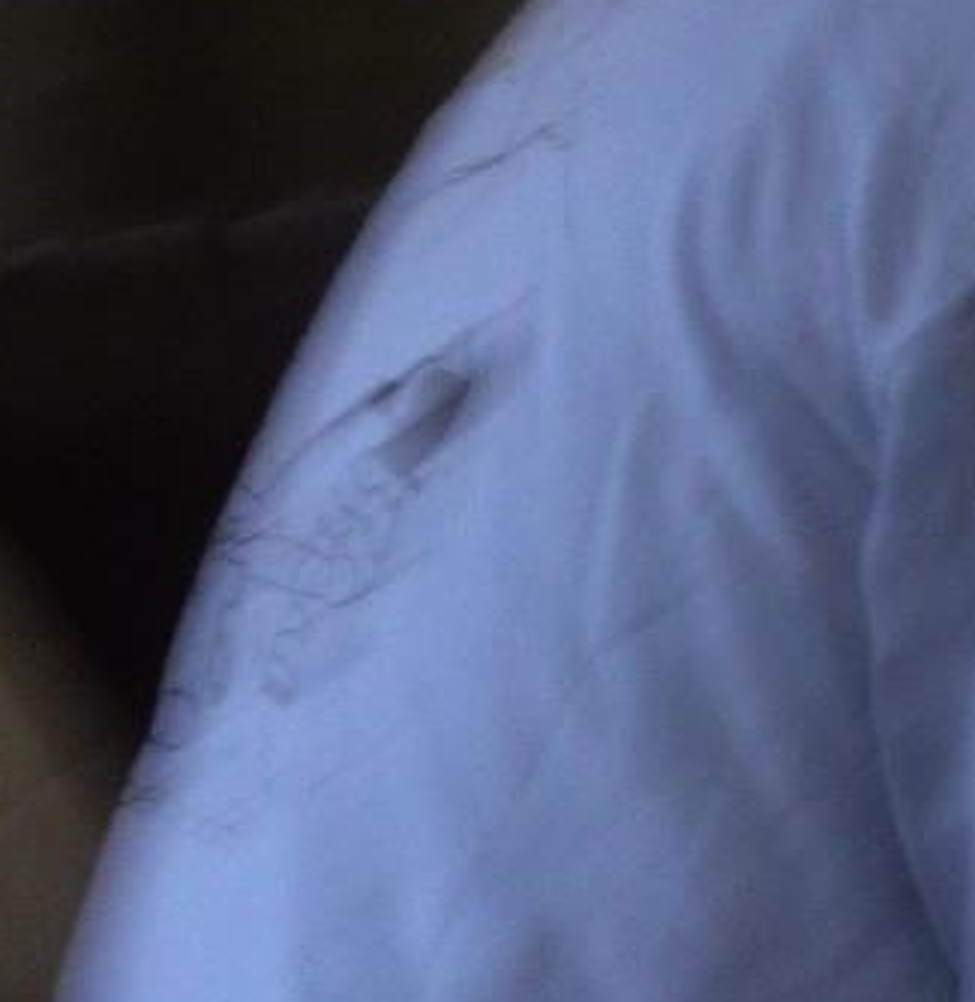} &
\includegraphics[width=0.16\textwidth]{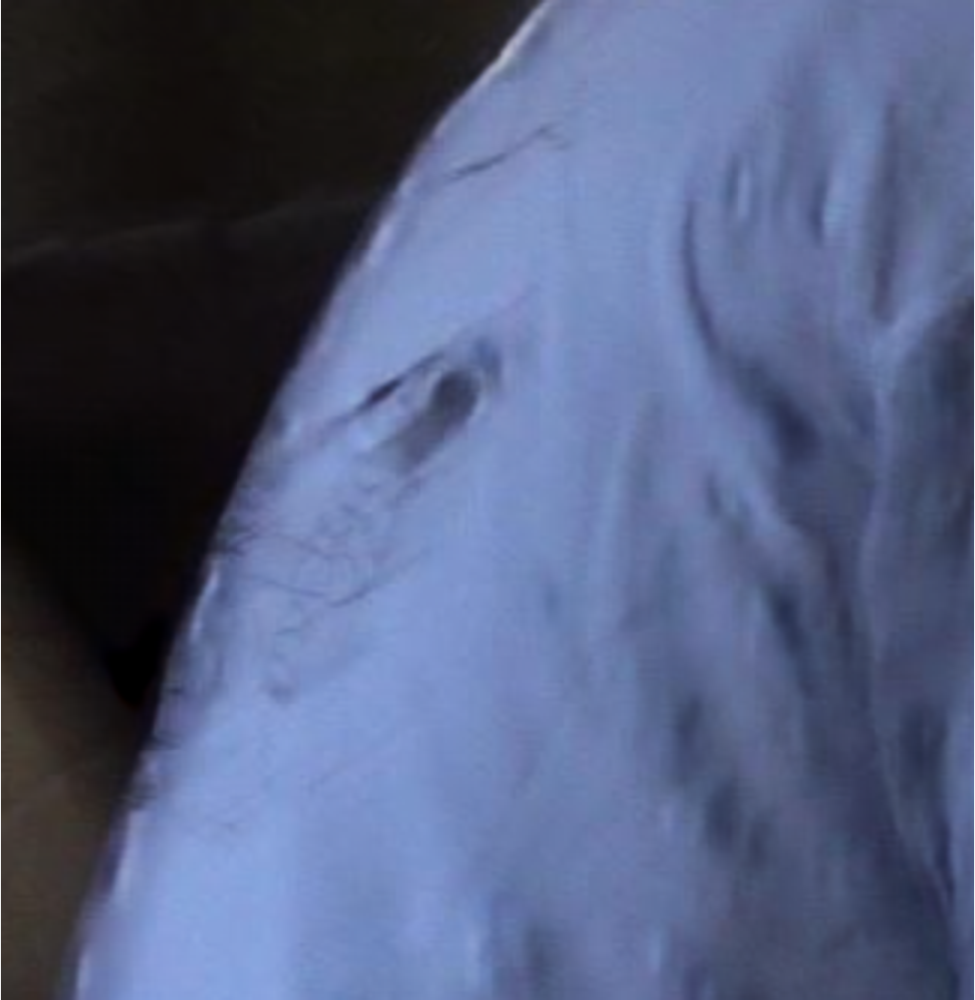} &
\includegraphics[width=0.16\textwidth]{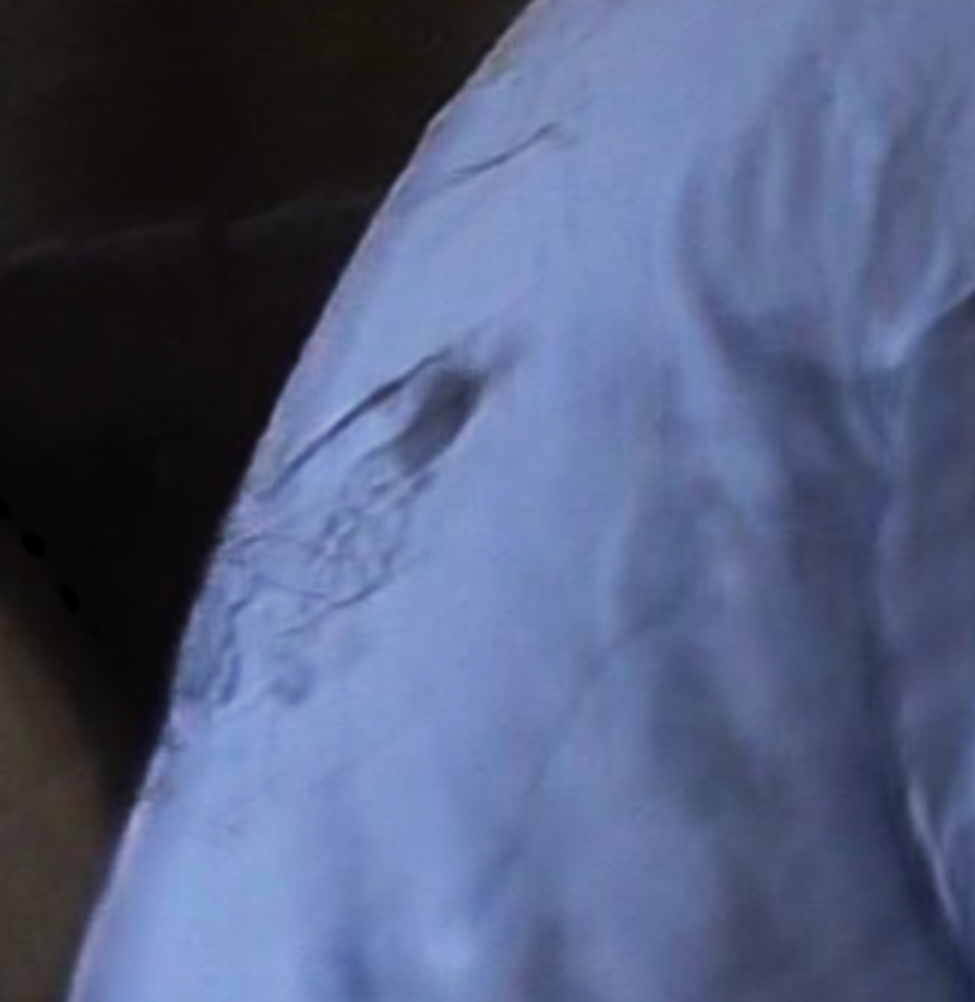} &
\includegraphics[width=0.16\textwidth]{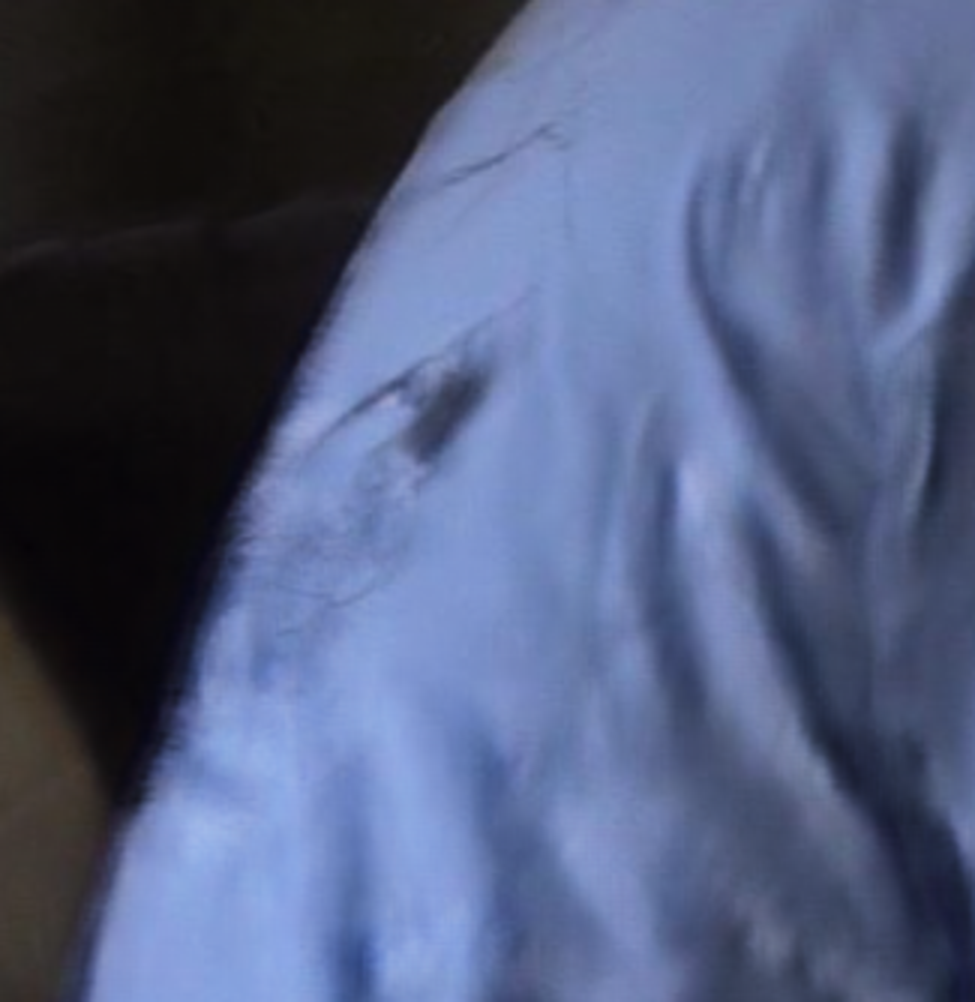} &
\includegraphics[width=0.16\textwidth]{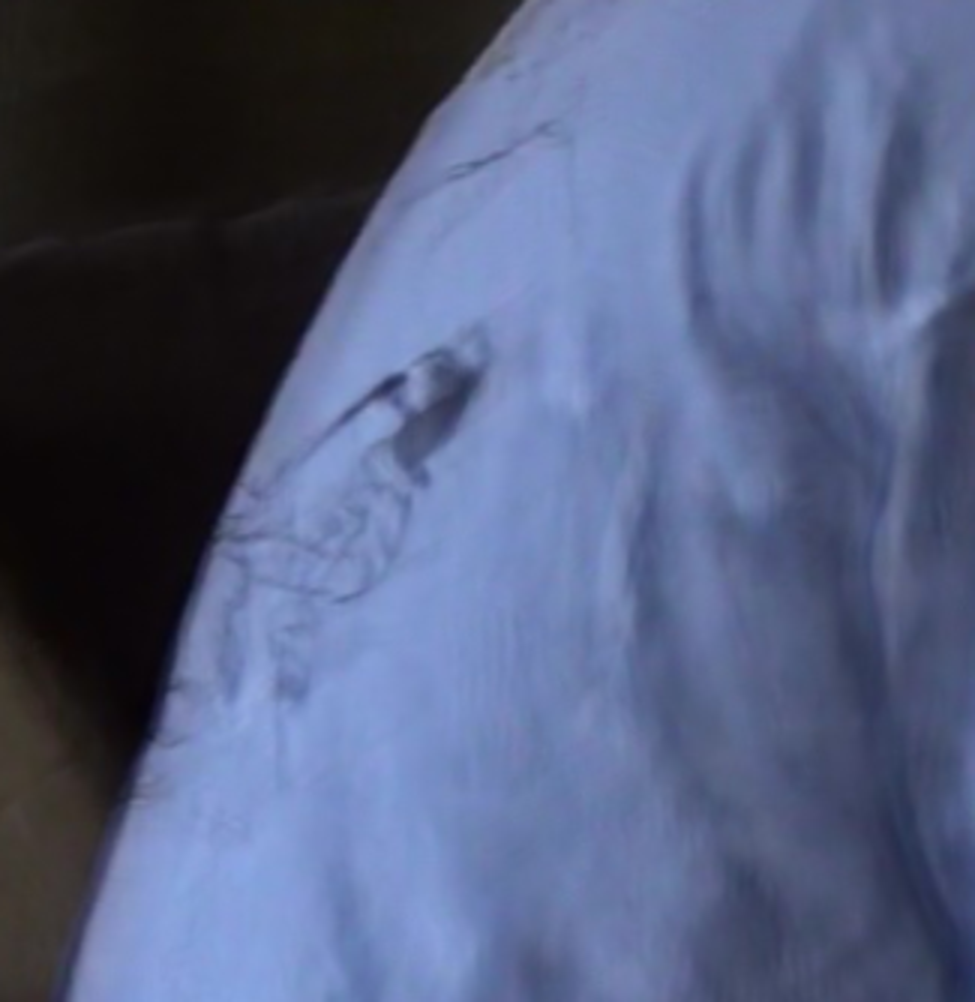} &
\includegraphics[width=0.16\textwidth]{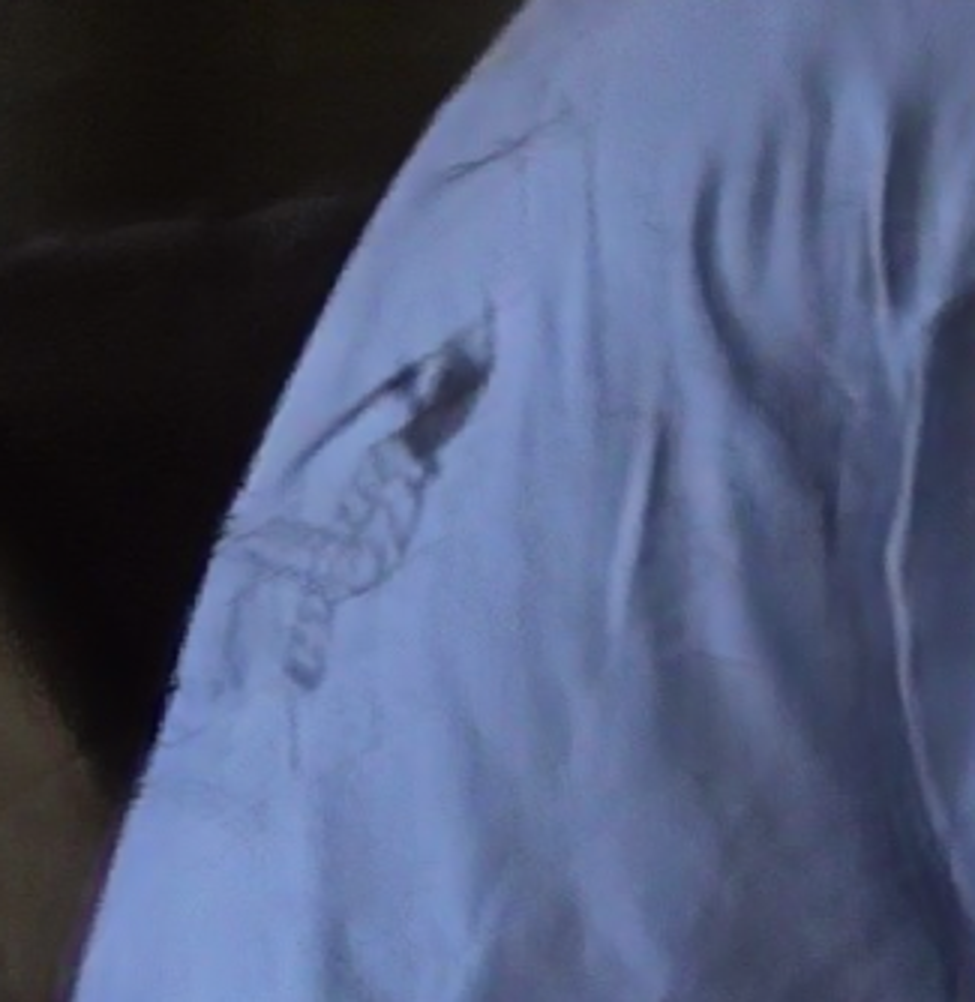} \\ [-1pt]
\end{tabular}
\setlength{\tabcolsep}{6pt}
\begin{tabular}{@{}llllll@{}}
(a) Blurry image & (b) DVD (single) & (c) DVD (noalign) & (d) DVD (flow) & (e) DAVID (DVD) & (f) DAVID (C-DVD) \\
\end{tabular}
\vspace{-0.5em}
\caption{Visual result comparison on real blurry videos. The second and fourth rows depict enlarged local areas of the first and third rows, respectively, where blur artifacts could be most easily observed and compared around the boundary of the front fender (second row), and the texture on clothes (fourth row).}
\label{fig:visual_alley}
\vspace{-2.5mm}
\end{figure*}

\subsection{Evaluation on the Challenging DVD Dataset}
The original DVD dataset only considers a fixed blur level, \ie, synthesized by averaging $7$ frames. We aim to highlight the capability of deblurring models in dealing with different levels of blur. In Challenging DVD dataset, we purposely synthesize the blur of multiple levels by averaging the frames with different window size such as $3$, $7$, $11$ or $15$. We also form a 10-fold testing split that has no data overlap according to \cite{su2017deep}. Our testing set consist of 297, 349, 739, 524 samples for C-DVD-3, C-DVD-7, C-DVD-11, C-DVD-15, respectively. We compare DAVID with two state-of-the-art video deblurring methods, WFA \cite{delbracio2015burst} and DVD\cite{su2017deep}; both are also fine-tuned on the Challenging DVD dataset.

In Table~\ref{Challenging_DVD}, PSNR is averaged over all frames for each video. We clearly see that all the methods show performance degradation compared to performance on original DVD set (see Table~\ref{Original_DVD}), suggesting that our newly synthesized dataset is indeed more challenging. However, DAVID shows consistently superior performance as the dual attention module can dynamically gather the temporal information and thus handle different levels of blur.

\begin{table*}[htp!]
\vspace{-4em}
\centering
\begin{tabular}{ c | c  c  c  c  c  c  c  c  c  c | c}
    \toprule
    Methods & 1 & 2 & 3 & 4 & 5 & 6 & 7 & 8 & 9 & 10 & Average \\
    \midrule
    PSDEBLUR & $24.42$ & $28.77$ & $25.15$ & $27.77$ & $22.02$ & $25.74$ & $26.11$ & $19.71$ & $26.48$ & $24.62$ & $25.08$ \\
	WFA\cite{delbracio2015burst} & $25.89$ & $32.33$ & $28.97$ & $28.36$ & $23.99$ & $31.09$ & $28.58$ & $24.78$ & $31.30$ & $28.20$	& $28.35$ \\
    DVD (single)\cite{su2017deep} & $25.75$	& $31.15$ & $29.30$ & $28.38$ & $23.63$ & $30.70$ & $29.23$ & $25.62$	& $31.92$ & $28.06$	& $28.37$ \\
    DVD (noalign)\cite{su2017deep} & $27.83$ & $33.11$ & $31.29$ & $29.73$ & $25.12$ & $32.52$ & $30.80$	& $27.28$ & $33.32$ & $29.51$ & $30.05$\\
    DVD (flow)\cite{su2017deep} & $28.31$ & $33.14$ & $30.92$ & $29.99$ & $25.58$ & $32.39$ & $30.56$	& $27.15$ & $32.95$	& $29.53$& $30.05$ \\
    IFI-RN\cite{nah2019recurrent} & - & - & - & - & - & - & - & - & - & - & $30.80$ \\
    Reblur2Deblur\cite{chen2018reblur2deblur} & - & - & - & - & - & - & - & - & - & - & $31.37$ \\
    \hline
    Ours (Single) & $29.95$ & $34.98$ & $33.12$ & $31.61$ & $26.09$ & $33.35$ & $31.45$ & $27.79$ & $35.66$ & $30.11$ & $31.41$ \\
    Ours (Internal Att) & $30.28$ & $35.20$ & $33.25$ & $31.81$ & $26.37$ & $33.47$ & $31.61$ & $28.00$ & $35.75$ & $30.37$ & $31.65$ \\
    Ours (External Att) & $30.38$ & $35.27$ & $33.33$ & $31.99$ & $26.45$ & $33.55$ & $31.64$ & $28.03$ & $35.79$ & $30.44$ & $31.71$ \\
    Ours (Dual Att) & $\textbf{30.68}$ & $\textbf{35.61}$ & $\textbf{33.59}$ & $\textbf{32.19}$ & $\textbf{26.78}$ & $\textbf{33.87}$ & $\textbf{31.96}$ & $\textbf{28.35}$ & $\textbf{36.14}$ & $\textbf{30.73}$ & $\textbf{31.99}$ \\
    \bottomrule
  \end{tabular}
  \vspace{-0.5em}
\caption{PSNR comparison on the Original DVD dataset. Best results are shown in bold.}
\label{Original_DVD}
\vspace{-5mm}
\end{table*}

\subsection{Ablation Study on Challenging DVD}
\subsubsection{Effect on Different Model Components}
\vspace{-0.5em}
In Table~\ref{Challenging_DVD}, under the same number of averaged frames, \ie, On C-DVD-11, we consistently observe that single backbone models are always sub-optimal compared to the one with internal attention module, indicating the effectiveness to aggregate the temporal information across the consecutive video frames. Meanwhile, Table~\ref{Challenging_DVD} shows that with the same setting, \ie all single backbone setting or all with internal attention setting, the larger the window size of averaging, the better the performance is, which suggests that larger window size is better for information gathering to conduct deblur. Moreover, our Dual Attention model further outperforms any internal attention model. This is achieved by further adding an external attention module, which aims at gathering the spatial correlation across different internal attention models.

\vspace{-0.5em}
\subsubsection{Effect on Stacked Frames}
\vspace{-0.5em}
We observe in Table~\ref{Challenging_DVD} that larger temporal window size is always beneficial than smaller window size. This motivates us to investigate how the temporal scale influences the overall performance. We fix a single backbone model, \ie, single backbone trained on C-DVD-11, and prepare the testing data by stacking frames of 1, 3, 5, 7 and 9 respectively. Table~\ref{ablation_depth} shows the ablative numbers with respect to the number of stacked frames. As the stacking number increases, we observe the performance increases. The larger window size indeed provide more temporal information for deblurring. However, when the number continues to increase, we see the performance is saturated. We observe the similar trend for other single backbone models trained on C-DVD-3 and C-DVD-7. It is because too large window size does not provide more information as two frames far apart might be irrelevant. Thus, we empirically find the optimal number of stacked frames (7 in our experimental setting) and apply it to conduct all our experiments.
%
%
%

\vspace{-0.5em}
\subsection{Attention Map Visualization}
\vspace{-0.5em}
To better understand the mechanism of our dual attention model, we visualize the attention maps given an input frame in Figure~\ref{fig:attention_map}. The frame is from the real blurry video ``piano'' in the qualitative testing set provided by \cite{su2017deep}. Inside the sample frame, there are multiple blur sources, including hand-held camera shake (which have caused global scene movement in video), and the local motion blurs caused by the player's head and hand moment. In this specific case, the camera movement is much stronger than object movement.

Figure~\ref{fig:attention_map} (3)-(5) show the external attention maps for Internal Att 3, Internal Att 7 and Internal Att 11 respectively. All three emphasize different detail structural information, which suggests that the external attention modules \textbf{do not degenerate} into selecting one out of the three. 
Notice that the third map shows the largest response magnitude, indicating that the external attention module favors more on the averaging 11 frames channel. We further acquire that the blur caused by global hand-held movement is closest to the blur by averaging 11 frames.
Such selectivity also shows to be spatially variant, which aligns with our hypothesis. For example, the third channel (averaging 11 frames) seems to account for most global blurs (in the majority area of the non-moving background); meanwhile, the first channel (averaging 3 frames) captures more responses in the area of the player's head and hands, which are moving objects.

Each group of internal attention maps then tend to further decompose the blur information into finer scales that possess multiple-scale temporal correlations. (6)-(9) show mostly spatial low frequency response, which indicates that the backbone branches trained on averaging $3$ frames input captures less motion. In contrast, (10)-(13) and (14)-(17) show response maps with plenty of the edge and region structural information,
indicating that the backbone branches trained on averaging $7$ and $11$ frames capture most motion and blur information for this specific video.

\subsection{Evaluation on Original DVD Dataset}
\vspace{-0.5em}
Original DVD dataset provides training data synthesized only by averaging $7$ frames. It is not most suited for the temporal-spatially varying blur case that DAVID targets. However, as a general public benchmark, we report PSNR performance on DVD to provide a reference comparison.
As shown in Table~\ref{Original_DVD}, our single backbone already outperforms the state-of-the-art Reblur2Deblur model \cite{chen2018reblur2deblur}. It also surpasses \cite{su2017deep} by 2dB, and more than 3.5dB over \cite{delbracio2015burst}. Further equipped with the proposed attention modules, our DAVID model yields stronger PNSR results with a 0.58dB gain over the single backbone. Notice that the gain is understandably smaller compared to the Challenging DVD case, since the blur here is not variant by averaging merely 7 frames. 

\subsection{Qualitative Results on Real Blurry Videos}
\vspace{-0.5em}
Besides the evaluation on synthetic data, we visually illustrate the effectiveness of our DAVID model on the real videos. Figure~\ref{fig:visual_alley} shows two examples from the qualitative testing set provided in \cite{su2017deep}. We compare three versions of author-provided DVD models: single, no-align and flow; and two DAVID versions: one trained on the original DVD dataset, the other on the Challenging DVD (C-DVD) dataset. 

As shown from the input zoom-in crop region, real videos have rich heterogeneous blurs at local regions. We observe that results from DVD (single) present significant ringing artifact, \ie on the front tire cover of the motorcycle and on the shirt's edge. DVD (no-align) alleviates but still with certain artifacts alongside the shirt's edge and the front fender. DVD (flow) for the first example still leaves unattended blurs while for the second example produces extra fake wrinkle at the bottom of the shirt. In contrast, DAVID (DVD) and DAVID (C-DVD) shows far fewer artifacts while preserving the sharp structural information. 

Moreover, when comparing DAVID (DVD) to DAVID (C-DVD), we still observe some slight artifact for the former, \ie, blurry boundary region under the light of the first example. 
Therefore with the same proposed model, training on more complicated synthetic video blur can help real-world generalization, which justifies the necessity of the proposed Challenging DVD dataset.

\subsection{Discussion on Temporal dependency}
\vspace{-0.5em}
We use sets of raw frames and average them to synthesize the blurry central frames, which has no overlap with each other. However, these sets still share temporal content dependency, for which we consider the consistency across a wider range of temporal scales, \eg consistent motion of foreground objects, which is commonly exploited in video restoration literature~\cite{liu2017robust}. Thus, we consider the temporal dependency as how many of the ``sets'' could be involved for restoring one central frame, rather than exposure period of a single set. Exploiting temporal dependency means adaptive and selective fusion of those different scales by internal attention. 
Certainly, central and neighboring frames are important. But wider temporal dependency is not negligible. In Table~\ref{ablation_depth}, when stacking more blurred frames (more ``sets''), the performance improves up to 7 stacking frames. Our other empirical results also endorses the importance of exploiting longer temporal dependency.

%


%% file: conclusion.tex
\vspace{-1em}
\section{Conclusions}
\vspace{-0.5em}
This work proposed the DAVID framework for blind video deblurring. The internal attention model is trained to adaptively select the temporal scales, while the external attention model further spatially aggregates the output from each internal module at the pixel-level. Different from previous synthetic data focusing on only a single blur level, we propose a Challenging DVD dataset (C-DVD) which incorporates multiple levels of blur by pooling video frames with different window sizes. Experiments on both datasets demonstrate that DAVID achieves better PSNR performance compared to several state-of-the-arts. Qualitatively, we further validate the advantage of our method in recovering sharp appearance with fewer artifacts, on real blurry videos. Ablation study shows that the dual attentions mutually benefit with learned structurally meaningful attention maps. Our future work will develop the unsupervised adaption of DAVID to real-world blurry videos. 